\documentclass{article}

\PassOptionsToPackage{numbers, compress}{natbib}
\usepackage[main, final]{neurips_2026}
\usepackage{amsmath}
\usepackage{adjustbox}
\usepackage{subcaption}
\usepackage[utf8]{inputenc} 
\usepackage[T1]{fontenc}    
\usepackage{hyperref}       
\usepackage{url}            
\usepackage{booktabs}       
\usepackage{amsfonts}       
\usepackage{nicefrac}       
\usepackage{microtype}      
\usepackage{xcolor}         
\usepackage{graphicx}
\usepackage{booktabs}
\usepackage{multirow}
\usepackage{wrapfig}
\usepackage{xcolor}
\usepackage{tabularx}
\usepackage{booktabs}
\usepackage{array}
\usepackage[table]{xcolor}
\definecolor{badred}{RGB}{180,45,45}
\definecolor{goodgreen}{RGB}{35,130,80}
\definecolor{softred}{RGB}{255,245,245}
\definecolor{softgreen}{RGB}{242,252,247}
\definecolor{softgray}{RGB}{248,248,248}
\usepackage[most]{tcolorbox}
\usepackage{xcolor}

\definecolor{boxgray}{RGB}{68,68,68}
\definecolor{lightboxgray}{RGB}{247,247,247}

\newtcolorbox{promptbox}[1]{
  enhanced,
  breakable,
  colback=lightboxgray,
  colframe=boxgray,
  colbacktitle=boxgray,
  coltitle=white,
  title={#1},
  fonttitle=\bfseries,
  boxrule=1.1pt,
  arc=4pt,
  left=10pt,
  right=10pt,
  top=10pt,
  bottom=10pt,
  toptitle=4pt,
  bottomtitle=4pt,
  lefttitle=10pt,
  righttitle=10pt,
  before skip=10pt,
  after skip=10pt
}
\title{Fisher-R1: Training LLM Agents for Reliable Hypothesis Testing}

\author{%
  Jiacheng Miao$^{1,2,*}$ \quad
  Jin Mu$^{3,*}$ \quad
  Guanhua Chen$^{3,4}$ \quad
  James Zou$^{1,5,6}$ \\[0.5em]
  $^{1}$Department of Biomedical Data Science, Stanford University \\
  $^{2}$Department of Genetics, Stanford University \\
  $^{3}$Department of Biostatistics and Medical Informatics, University of Wisconsin--Madison \\
  $^{4}$Department of Statistics, University of Wisconsin--Madison \\
  $^{5}$Department of Electrical Engineering, Stanford University \\
  $^{6}$Department of Computer Science, Stanford University \\[0.3em]
  $^{*}$Equal contribution
}
\begin{document}

\maketitle

\begin{abstract}
Reliable hypothesis testing is the foundation of many empirical scientific claims. Large language model (LLM) agents are increasingly used to automate this process, as they can inspect datasets, generate code, and produce analyses end-to-end. However, we show that they frequently make subtle inferential errors that lead to incorrect conclusions despite correctly executed analyses. Existing benchmarks fail to capture this failure mode, as they rarely assess whether a reported p-value is statistically valid given the assumptions underlying the data. We address this gap by building \textbf{P-Bench}, a benchmark comprising 425 open-ended, realistic hypothesis-testing tasks spanning economics, biology, and medicine. Each task requires an agent to select a statistical method, compute a p-value, and draw a conclusion given only a scientific hypothesis and a dataset. We further introduce \textbf{Fisher-R1}, an open-weight LLM agent trained for rigorous hypothesis testing using synthetic tasks and reinforcement learning. On P-Bench, Fisher-R1-14B substantially improves over its backbone and \textbf{outperforms strong proprietary and open-source baselines, including GPT-5.4 and DeepSeek-V4-Pro}, achieving a 21\% average relative improvement in single-trial success over DeepSeek-V4-Pro, with gains up to 26\% on the most challenging tasks.  Our results demonstrate that current LLM agents lack reliable statistical reasoning for hypothesis testing and that reinforcement learning on tasks with verified statistical reward substantially improves reliability. Code is available at \url{https://github.com/jmu27/FisherR1}.

\end{abstract}

\section{Introduction}

\begin{figure}[h]
  \centering
\includegraphics[width=0.95\textwidth]{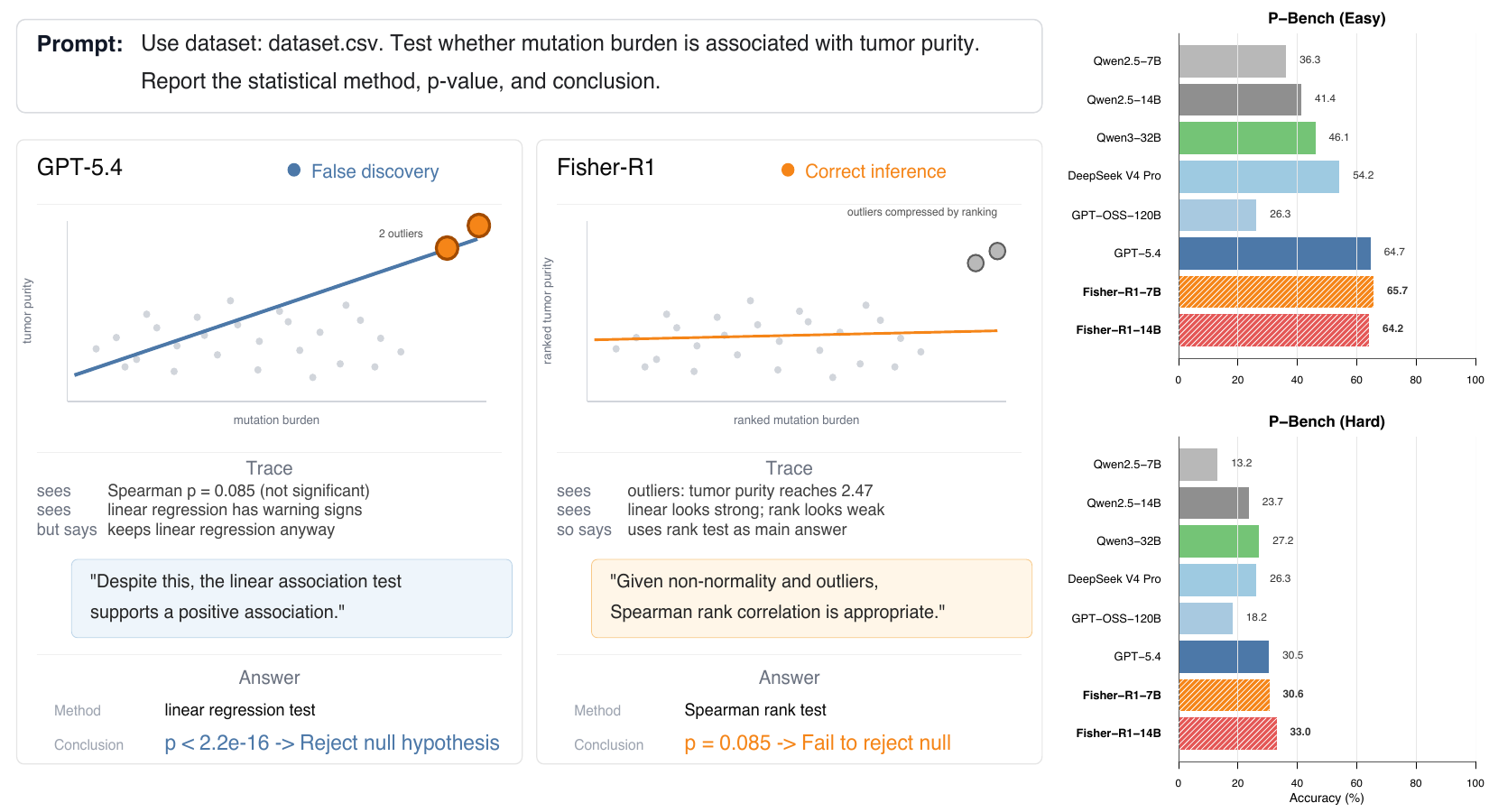}
  \caption{\textbf{Open-ended hypothesis testing with coding agents.}
 Left: GPT-5.4 identifies outliers but still applies linear regression, leading to a false discovery. Fisher-R1 instead uses a rank-based test and correctly fails to reject the null. Right: P-Bench accuracy on easy and hard splits.}
  \label{fig1}
\end{figure}

Many areas of empirical science follow a common workflow: a researcher begins with a scientific question and a dataset, translates the question into a testable hypothesis, selects an appropriate statistical test, computes a p-value, and then draws a conclusion \citep{Wasserstein02042016}. This hypothesis-testing framework appears in high-stakes settings across disciplines, from medicine to economics. For example, in clinical trials, researchers test whether a new treatment is more effective than a placebo, while in economics, policymakers evaluate whether interventions such as minimum wage increases affect employment outcomes. The method must match the question and data; otherwise, a precise p-value can support the wrong scientific conclusion.

Large language model (LLM) coding agents can now execute much of the empirical workflow: inspecting datasets, generating code, running analyses, and drafting reports \citep{zhang2023data, zhang2025deepanalyze,hong2025data}. However, hypothesis testing is not merely code execution. Consider a cancer genomics task testing whether mutation burden is associated with tumor purity. The dataset contains a few high-leverage outliers: a linear analysis reports a highly significant association driven by these points, whereas a rank-based test does not. As shown in Figure \ref{fig1}, a frontier LLM agent recognizes these warning signs yet proceeds with linear regression, producing runnable code and a fluent but incorrect conclusion. In contrast, a Spearman test correctly fails to reject the null. 

This example highlights how LLM agents can generate technically valid analyses while making flawed inferential choices that lead to false discoveries. Existing evaluations do not directly measure this ability. General data-analysis benchmarks \citep{hu2024infiagentdabenchevaluatingagentsdata, egg2025dabstep,huang-etal-2024-da,zhu2024large} often reward a plausible answer or an executable workflow, but they rarely isolate the inferential method itself: which test was executed, whether the reported p-value is grounded in that execution, and whether the conclusion follows from the computed evidence. The gap is especially consequential for open-ended scientific discovery, where agents may search many datasets and hypotheses, and for private scientific and medical data, where reliable local or open-weight agents are needed.

Closing this evaluation gap requires verified statistical tasks: a scientific question, a dataset, an executable analysis, and a checked answer key for the method, p-value, and conclusion. Such tasks are scarce. Papers report selected results, public datasets require cleaning, and statistical methods are not packaged as machine-checkable targets.

To address this gap, we introduce \textbf{P-Bench} (\textbf{P-value Bench}), a benchmark of 425 open-ended hypothesis-testing tasks built on real scientific data. Each task gives an agent a scientific question, a dataset, and a data description. The agent must choose a statistical method, execute the analysis, report a p-value, and draw a conclusion. Tasks are drawn from peer-reviewed economics and biology papers and from authoritative biostatistics teaching materials. Every answer key is grounded in a logged execution of the canonical reference analysis and audited by domain experts. To make this construction tractable, we use an automated reproducibility pipeline that links each published claim to its dataset and analysis code. The pipeline is described in the appendix.

We further train \textbf{Fisher-R1}, an open-weight LLM agent for rigorous hypothesis testing. To enable training at scale, we design a synthetic task generator that produces realistic hypothesis-testing scenarios with verified answer keys. Fisher-R1 is trained on these tasks with supervised fine-tuning followed by reinforcement learning. On P-Bench, Fisher-R1-14B outperforms strong proprietary and open-source baselines, including GPT-5.4 and DeepSeek-V4-Pro.

To summarize, our contributions are:

\begin{itemize}
 \item We introduce \textbf{P-Bench}, an expert-verified, executable benchmark of 425 open-ended hypothesis-testing tasks on real scientific data, evaluating LLM agents on p-value accuracy and conclusion correctness.
\item We train \textbf{Fisher-R1}, an open-weight hypothesis-testing agent, using synthetic executable tasks and outcome-grounded reinforcement learning. Fisher-R1 substantially improves over its backbone and outperforms GPT-5.4 and DeepSeek-V4-Pro on P-Bench.
\end{itemize}

\section{Problem Formulation}
\subsection{Settings}
We consider open-ended hypothesis testing with LLM coding agents. Each task gives the agent a scientific question $q$, a dataset $D$, and a data description $s$, but not the statistical method. Given $x=(q,D,s)$, the agent must choose a method, execute the analysis, return a p-value $\hat{p}$, and a decision $\hat{\delta}$, where $\hat{\delta}$ indicates whether to reject the null hypothesis at a pre-specified significance level.

Evaluation compares the agent output with a hidden answer key $k^\star=(p^\star,\delta^\star)$. The agent interacts with an R environment through a multi-turn loop, reflecting the statistical software ecosystem used in many empirical analyses and providing a fixed execution backend for evaluation. At step $t$, the history is
$h_t=(x,a_1,o_1,\ldots,a_{t-1},o_{t-1})$,
where $a_i$ is an agent action and $o_i$ is the corresponding environment observation. An action may contain R code, an intermediate analysis decision, or the final answer. The environment executes code actions and returns observations such as data summaries, warnings, model outputs, diagnostics, test statistics, and p-values. The completed trajectory is
\[
\tau \sim \pi_\theta(\cdot \mid x),
\]
where $\tau$ includes both the executed code trace and the final report. At evaluation time, we parse $\tau$ to compare the reported p-value and the final decision with $k^\star$.

\subsection{Related works}
\paragraph{LLM agents for data analysis and statistical reasoning.}
Recent work has evaluated LLM agents that write code, inspect data, execute programs, and iteratively revise their solutions, building on the ReAct paradigm of interleaved reasoning and tool use \citep{yao2022react}. A substantial body of data-analysis benchmarks evaluates whether such agents can answer factoid queries about a dataset (e.g., ``what is the mean of $X$?'') or build predictive models on tabular data through executable workflows \citep{lai2023ds,hu2024infiagent,egg2025dabstep,huang-etal-2024-da,zhang2025datascibench,chan2024mle,li2024tapilot,li2025ida}. These benchmarks score the executability of the workflow and the exact-match correctness of a final factoid or model accuracy, but they neither require nor verify that the chosen statistical procedure is valid for the data at hand. A complementary line, exemplified by StatQA, asks whether LLMs can select an appropriate statistical procedure in a multiple-choice format, without executing the analysis with code \citep{zhu2024large}. P-Bench targets a regime that neither captures the full open-ended hypothesis-testing loop, method selection, executed analysis, p-value, nor the reject/fail-to-reject decision, and is evaluated against an answer key derived from the canonical reference analysis on real scientific data.

\paragraph{Scientific claim verification and reproducibility.}
Scientific claim verification has been studied as an evidence-retrieval and textual-entailment problem: SciFact asks models to verify scientific claims against evidence-containing abstracts and supply supporting rationales \citep{wadden2020fact}, with follow-on benchmarks scaling this to larger and multimodal corpora \citep{kumar2025sciclaimhunt,lal2025musciclaims}. This line evaluates whether claims are supported or refuted by the literature, but does not reconstruct the underlying statistical analysis. Large-scale replication efforts such as SCORE further show that re-running and checking empirical claims is important but costly, requiring substantial human effort to assess robustness and test replicability \citep{miske2026investigating}. These efforts target manuscript-level verification of whether published numbers can be reproduced. In contrast, P-Bench evaluates the upstream act of inference itself: given the same data and question, does the agent select a method that yields a statistically valid p-value?

\paragraph{Autonomous research agents.}
A recent line of work pursues end-to-end autonomous scientific discovery, with LLM agents that propose hypotheses, generate code, run experiments, and draft manuscripts \citep{lu2026towards,tang2025ai,yamada2025ai}. The reliability of such pipelines is, however, bounded by the statistical reasoning of the underlying agent at every stage. As our P-Bench experiments show, current LLM agents may make subtle inferential errors even on well-defined hypothesis-testing tasks, undermining the validity of any conclusion drawn downstream and casting doubt on whether fully autonomous end-to-end discovery is presently achievable. Fisher-R1 directly targets this bottleneck.

\section{P-Bench}
\label{sec:pbench}
P-Bench is a benchmark of open-ended hypothesis-testing tasks built from real scientific data spanning economics, biology, and medicine. Each task is anchored to a statistical claim from a high-quality source, such as peer-reviewed top scientific journals and canonical course notes, so the answer key reflects a hypothesis test that domain experts have already computed and acted on.

\begin{figure}[!h]
  \centering
    \includegraphics[width=\textwidth]{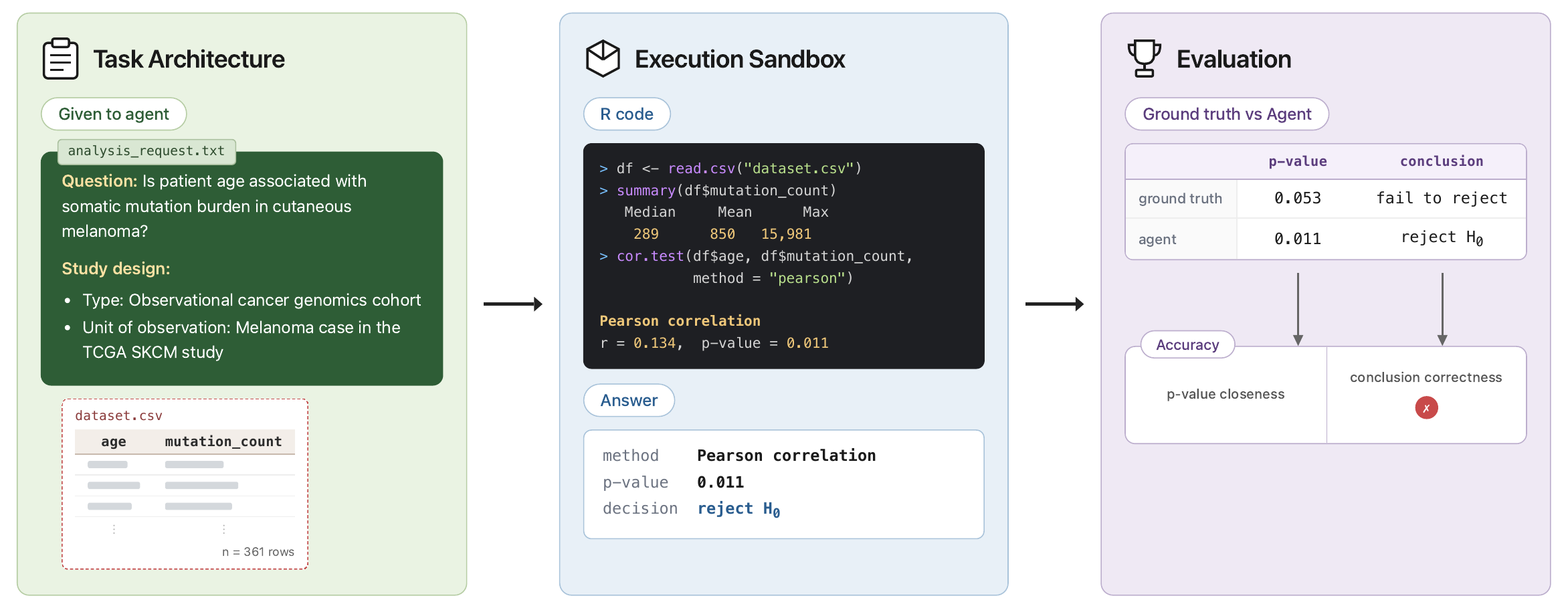}
  \caption{A P-Bench task is composed of an analysis request, a CSV dataset, and a hidden answer key derived from the canonical reference analysis. The agent writes and executes R code inside a logged sandbox and returns a chosen statistical method, $p$-value, and reject/fail-to-reject decision, which are then compared against the answer key.}
  \label{fig2}
\end{figure}

\subsection{Task Formulation}

A P-Bench task provides the agent with (Figure~\ref{fig2}): (i) an analysis request describing a scientific question and the associated data, and (ii) a dataset in CSV format. The agent must explore the data, write and execute R code in a logged sandbox environment, and return a structured output consisting of a selected statistical method, a $p$-value, and a reject/fail-to-reject decision. Importantly, the task does not prescribe a statistical method. Instead, the agent must autonomously select an appropriate analysis strategy, and its output is evaluated against a hidden answer key comprising the $p$-value, and a decision derived from the canonical analysis.

\subsection{Dataset Construction and Verification}

P-Bench is built from three high-quality source families: economics papers in top journals with datasets hosted on Harvard Dataverse \citep{dfeep_harvard_dataverse}, peer-reviewed biology papers with datasets hosted on cBioPortal\citep{cerami2012cbio,gao2013integrative,wang2025biodsa1kbenchmarkingdatascience}, and authoritative teaching materials from Vanderbilt Biostatistics\citep{harrell_hbiostat} with datasets distributed through R packages. Each task is derived from a real scientific analysis originally conducted by domain experts. To ensure diversity and balance, we first collected a broad pool of candidate tasks and randomly sampled from them to construct an evenly distributed benchmark. We also introduced controlled perturbations to the datasets to better reflect the noise, inconsistencies, and irregularities commonly encountered in real-world data analysis.

Starting from statistical claims extracted from the sources, we applied a three-stage pipeline: (1) reproduce the reference analysis on a clean machine and log the execution; (2) filter out analyses that could not be reproduced, and ground each remaining claim in the execution log; and (3) generate a self-contained executable task paired with a structured answer key in the P-Bench format. As a result, every released task is reproducible, traceable to an executed reference run, and self-contained.

P-Bench tasks and their answer keys are trustworthy by construction: keys are computed from canonical reference code (not transcribed from paper text), cross-validated against the published claim, and every task is audited by domain experts. (i) Reproducible execution. Every $p$-value, test statistic, and decision is read off a logged run of canonical reference code: a deposited replication script for economics, a methods-section re-implementation on the released tables for biology, or the published course-notes analysis for biostatistics. (ii) Expert review. Reviewers trained in statistics and biostatistics independently audit each finalized task, verifying that the analysis request, released data subset, and answer key all align with the source; tasks failing review are repaired or removed.

\subsection{Composition}

After verification, P-Bench contains \textbf{425} high-quality open-ended hypothesis-testing tasks, stratified into Easy (203) and Hard (222) splits using deterministic, metadata-only rules: a task is Hard if it belongs to a reference-method family that requires careful model specification and assumption checking (e.g., Cox regression, IV/2SLS, Tobit model) or it carries an adversarial data-quality perturbation. Several properties make P-Bench distinctive:

\textbf{Diverse and realistic methods.} P-Bench covers \textbf{17} hypothesis-testing method categories spanning commonly used tests ($t$-test, $\chi^2$, OLS coefficient test, Fisher exact test) and specialized methods routinely used in real published analyses (e.g., Cox proportional-hazards regression, IV / 2SLS estimators, mixed-effects models, log-rank tests, Mann--Whitney, Tobit model, probit model). \begin{wrapfigure}{r}{0.5\textwidth}
    \centering
    \vspace{-0.5em}
    \includegraphics[width=0.5\textwidth]{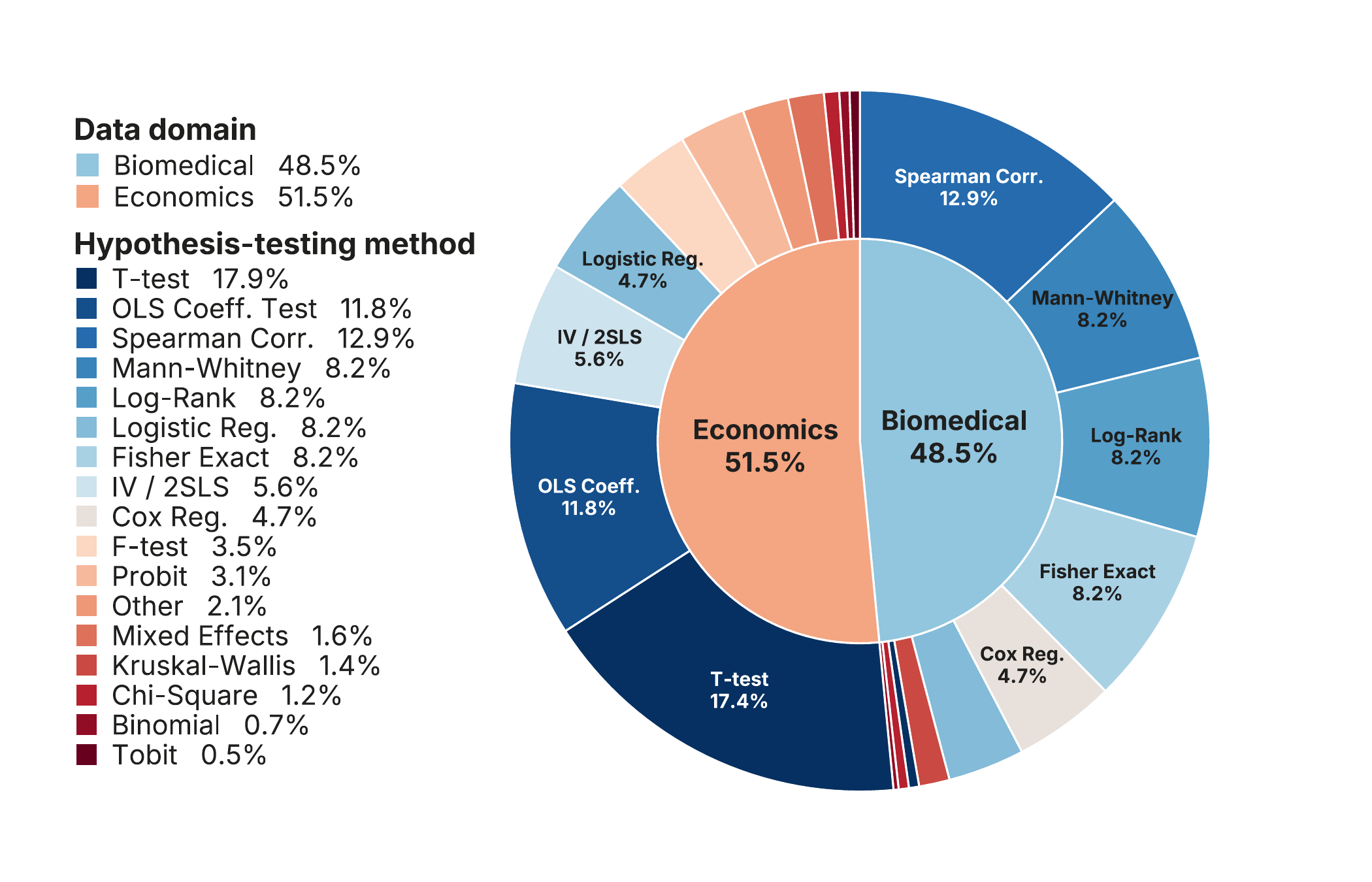}
    \caption{Task Composition of P-Bench}
    \label{fig:Composition}
    \vspace{-1em}
\end{wrapfigure} No single category exceeds 19\%, so an agent cannot succeed by defaulting to a fixed recipe.

\textbf{Cross-domain coverage.} Tasks span economics, biology, and medicine, covering both randomized experiments and observational studies. Each setting has its own modeling assumptions and inferential targets, so an agent must transfer statistical reasoning across disciplines.

\textbf{Realistic statistical traps.} P-Bench contains adversarial data-quality and assumption-checking traps, including outliers, heteroskedasticity, and clustered observations, where naive methods yield confident-but-wrong results. These are textbook assumption violations that arise routinely in real-world data and that any trained statistician is expected to detect and handle.

\section{Fisher-R1 Training}
\begin{figure}[!h]
  \centering
    \includegraphics[width=\textwidth]{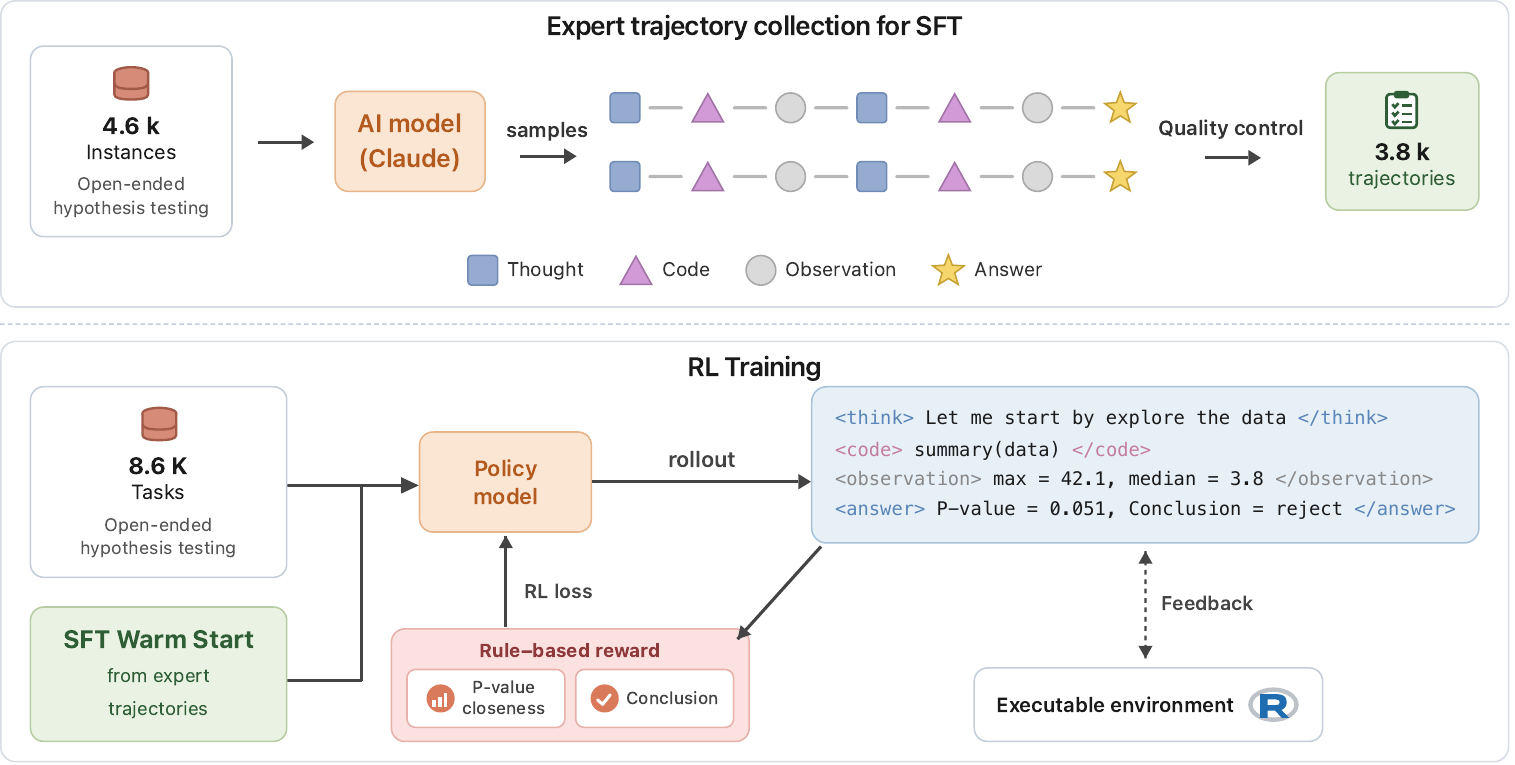}
  \caption{\textbf{Top}: Expert trajectories are collected for SFT. The model generates reasoning traces containing thoughts, code, observations, and final answers; quality control filters for high-quality trajectories. \textbf{Bottom}: The policy is initialized from the SFT model and optimized with RL. During rollout, it interacts with an executable environment to generate trajectories. Rule-based rewards score p-value accuracy and conclusion correctness.}
  \label{fig3}
\end{figure}

\subsection{Synthetic Data}
We train on synthetic tasks with executable data-generating processes, known target analyses, and programmatically checkable answer keys. To generate each task, we use an LLM to write both the simulation code (which produces the dataset) and the accompanying analysis question. The simulation runs the canonical statistical method on its own simulated data, and the resulting $p$-value, decision, and method label form the answer key. This gives us a scalable, verified reward signal for RL, which is the bottleneck for training on real, published data. By construction, it also keeps P-Bench fully out of the training set, allowing us to test out-of-distribution generalization.

\paragraph{Task taxonomy.}
Each training task is generated by combining six independent factors: a statistical method ($M$), a domain scenario ($S_m$), a sample size ($N$), an effect size ($E$), a prompt style ($P$), and a random seed ($K$). Formally, the task taxonomy is the Cartesian grid
$$
\underbrace{M}_{\text{method}}\times
\underbrace{S_m}_{\text{scenario}}\times
\underbrace{N}_{\text{sample size}}\times
\underbrace{E}_{\text{effect size}}\times
\underbrace{P}_{\text{prompt style}}\times
\underbrace{K}_{\text{seed}},
$$
Each method is paired with multiple domain scenarios that wrap a shared data-generating process in different realistic study descriptions. Effect size takes one of three regimes (null, borderline, medium); the borderline regime is auto-calibrated by simulation, so tasks fall in a genuinely ambiguous significance range. Prompt style controls whether the question states only the research question (\texttt{non-hint}) or additionally flags a methodologically relevant feature of the data, such as endogeneity or clustering (\texttt{hint}). Sample size and seeds are varied for coverage and reproducibility. Full per-axis level lists, the end-to-end generation pipeline, and example tasks are given in Appendix~\ref{app:synthetic-data}.

On top of the clean grid, we add data-quality perturbations such as missing values and extreme or invalid observations. These test whether the agent notices and handles problematic data instead of applying an analysis mechanically. The full corpus contains 8,642 tasks with balanced coverage.

\subsection{Supervised Fine-tuning}
We warm-start the policy model with supervised fine-tuning on teacher-generated analysis trajectories. The teacher trajectories are generated by Claude-Sonnet-4.6 \citep{anthropic2026claudesonnet46} on a randomly sampled subset of synthetic tasks, yielding 4,611 trajectories in total. For every task, the teacher follows a fixed five-step workflow: (i) basic exploratory data analysis (EDA) (data shape, column types, variable roles); (ii) detailed EDA (distributions, missingness, balance, correlations); (iii) assumption checking tailored to the candidate method (e.g., normality, proportional hazards, instrument strength, parallel trends); (iv) method selection and analysis; and (v) conclusion.

We then apply automatic quality control. A trajectory is retained only if it produces a valid multi-turn trace and a parseable conclusion. It must also match the ground-truth reject/fail-to-reject decision. When a $p$-value is reported, it must agree with the answer key in significance at $\alpha=0.05$ and within one order of magnitude. We further remove teacher-only metadata and any answer-key leakage from training prompts and targets. After filtering, we keep $3851/4611 \approx 83.5\%$ of trajectories for SFT.

Each trajectory is formatted in ReAct (CodeAct) style \citep{yao2022react,wang2024executable}, with the same \texttt{<think>}, \texttt{<code>}, \texttt{<observation>}, \texttt{<answer>} tags as the RL environment. Only assistant turns are optimized; user prompts and tool observations are provided as context. Let
\(\mathcal{D}_{\mathrm{SFT}}=\{(x_i,y_i)\}_{i=1}^{n}\) denote the filtered trajectory set, where \(x_i\) is the task prompt and \(y_i=(y_{i,1},\ldots,y_{i,T_i})\) is the teacher
response sequence. We minimize the masked autoregressive negative log-likelihood
\[
\mathcal{L}_{\mathrm{SFT}}(\theta)
=
-\frac{1}{\sum_i\sum_t m_{i,t}}
\sum_{i=1}^{n}\sum_{t=1}^{T_i}
m_{i,t}
\log \pi_\theta(y_{i,t}\mid x_i,y_{i,<t}),
\]
where \(m_{i,t}=1\) for assistant tokens and \(0\) for prompt and observation tokens. 

\subsection{Reinforcement Learning}
\label{sec:train-rl}

\paragraph{Algorithm.}
Starting from the SFT-initialized policy~$\pi_\theta$, we use the Decoupled Clip and Dynamic Sampling Policy Optimization (DAPO) algorithm~\citep{yu2025dapo} . For each prompt \(q\), we sample a group of \(G\) multi-turn rollouts
\(\{o_i\}_{i=1}^{G} \sim \pi_{\theta_{\mathrm{old}}}(\cdot \mid q)\)
and score each rollout with the Fisher reward \(R(o_i)\). The group-relative
advantage is
\[
\hat A_i
=
\frac{R(o_i)-\mu_q}{\sigma_q+\epsilon},
\qquad
\mu_q=\frac{1}{G}\sum_{j=1}^G R(o_j),
\qquad
\sigma_q^2=\frac{1}{G}\sum_{j=1}^G (R(o_j)-\mu_q)^2 .
\]
For rollout \(o_i=(y_{i,1},\ldots,y_{i,T_i})\), where \(T_i=|o_i|\) denotes the number of generated assistant tokens in rollout \(o_i\) that are included in the policy-gradient loss, the token-level importance ratio is
\[
r_{i,t}(\theta)
=
\frac{
\pi_\theta(y_{i,t}\mid h_{i,t})
}{
\pi_{\theta_{\mathrm{old}}}(y_{i,t}\mid h_{i,t})
},
\]
where \(h_{i,t}\) denotes the full interaction history before token \(t\), including the prompt, previous assistant tokens, and environment observations. We maximize the DAPO objective
\[
J_{\mathrm{DAPO}}(\theta)
=
\mathbb{E}_{q,\{o_i\}}
\left[
\frac{1}{\sum_{i=1}^G T_i}
\sum_{i=1}^G
\sum_{t=1}^{T_i}
\min\left(
r_{i,t}(\theta)\hat A_i,\,
\operatorname{clip}
\left(
r_{i,t}(\theta),1-\epsilon_{\ell},1+\epsilon_h
\right)\hat A_i
\right)
\right],
\]
where we use the asymmetric clipping range
\((\epsilon_{\ell},\epsilon_h)=(0.20,0.28)\).
The asymmetric range, with \(\epsilon_h>\epsilon_{\ell}\), permits larger updates
for tokens from positive-advantage rollouts while retaining the lower clip to
control excessively large probability decreases. Following DAPO, we use
\emph{dynamic sampling}: groups satisfying
\(\sigma_q^2=0\)
are discarded and re-sampled. Thus, each
policy update is computed from groups with non-degenerate relative advantages.

\paragraph{Reward.}
For each rollout~$o$, we compute a scalar reward from two
outcome-grounded components, gated by a hard format constraint:
\[
R(o)
=
I_{\mathrm{valid}}(o)
\left(
w_p r_p(o)
+
w_c r_c(o)
\right),
\qquad
w_p+w_c=1 .
\]
We set \(w_p=0.9\), \(w_c=0.1\).  The conclusion reward is assigned a smaller weight because it primarily checks whether the reported decision is consistent with the reported \(p\)-value. Here \(I_{\mathrm{valid}}(o)=1\) only if the trajectory contains reasoning,
executable code, and a parseable terminal \texttt{<answer>...</answer>} block;
otherwise \(R(o)=0\). The \(p\)-value component compares the reported
\(\hat p\) with the answer-key value \(p^\star\) on a two-sided normal
test-statistic scale:
\[
z(p)=\Phi^{-1}(1-p/2),\qquad
r_p(o)=
\exp\!\left(
-\frac{|\min\{z(\hat p),5\}-\min\{z(p^\star),5\}|}{\sigma}
\right),
\quad
\sigma=1 .
\]
The map \(z(p)=\Phi^{-1}(1-p/2)\) does \emph{not} assume that the underlying
data are Gaussian; it is a monotonic, distribution-free re-labeling of the
\(p\)-value as the two-sided standard-normal \(z\)-score whose tail probability
equals \(p\), and applies regardless of which test (parametric or
nonparametric) produced \(p\). We score in \(z\)-space because raw
\(p\)-values are heavily compressed near \(0\), where the most informative
differences in evidence live. For example, the pair \(\{p=0.5, p=0.6\}\) and
the pair \(\{p=0.1, p=10^{-10}\}\) both have \(|\Delta p|\approx 0.1\), yet
the first represents essentially no change in evidence while the second spans
many orders of magnitude in statistical significance. The \(z\)-scale addresses this compression and is a widely used convention in statistical practice, such as meta-analysis. The conclusion component \(r_c\in\{0,1\}\) checks whether the final reject/fail-to-reject decision agrees with the answer key at \(\alpha=0.05\). We do not include an explicit method-correctness term in the reward because multiple statistical procedures may be defensible for the same hypothesis-testing task. Instead, method choice is evaluated indirectly through the outcome-grounded reward: an inappropriate method will typically produce a p-value that deviates from the reference analysis, and therefore receive a lower z-space p-value closeness reward.

\section{Experiment}
\subsection{Experimental Setting} 
\paragraph{Models and Baselines} We include Qwen2.5-Coder-7B and 14B \citep{hui2024qwen25codertechnicalreport} as our backbone models to compare different baselines. We compare Fisher-R1 with strong  proprietary models: GPT-5.4 \cite{openai2026gpt54}. We also include four outstanding open-source models: DeepSeek-V4-Pro \citep{deepseek2026v4}, GPT-OSS-120B \citep{openai2025gptoss}, Qwen3-Coder-30B \citep{qwen3coder30b} and Qwen3-32B \citep{qwen3}, and DataMind \citep{qiao2026scalinggeneralistdataanalyticagents} has been trained for data analysis. We further construct two baselines tailored to statistical analysis: StatQA-FT adapts StatQA's  supervised data and LoRA fine-tuning recipe \citep{zhu2024large}; DARE-retrieval  uses the released DARE retriever \citep{sun2026darealigningllmagents} to select the three most relevant R function documents from its R Programming Knowledge Base (RPKB).

\paragraph{Evaluation metrics.}
We evaluate each model on two splits of P-Bench: P-Easy and P-Hard. 
For each task, we sample three independent rollouts and report results under two correctness criteria. 
\textbf{Raw} measures whether the model reaches the correct hypothesis-testing conclusion at 
$\alpha=0.05$, i.e., whether its final reject/fail-to-reject decision matches the ground-truth decision. 
Unparseable rollouts are counted as incorrect. 
\textbf{Strict} further requires the reported $p$-value to be numerically close to the ground-truth value in two-sided $z$-space:
\[
\text{Strict}
=
\text{Raw}
\;\wedge\;
\left|z(\hat p)-z(p^\star)\right| < 0.5,
\qquad
z(p)=\Phi^{-1}(1-p/2).
\]
Thus, Raw evaluates conclusion-level correctness, while Strict evaluates whether the model reaches the correct conclusion with a sufficiently accurate reported $p$-value. For each model, we run three independent trials per task, pass@1 averages the per-trial success rate; pass@3 counts a task as solved if any of its three trials succeed. Because the canonical analyses come from peer-reviewed papers or authoritative course materials and are expert-audited, scoring is reproducible and expert-grounded.

\subsection{Main Results}


\begin{table}[h]
\centering
\caption{Main results on P-Bench. Raw denotes correct conclusion direction, while Strict additionally requires $|\Delta z| < 0.5$, where $z$ is the two-sided $z$-score derived from the reported $p$-value. pass@1 is reported as mean$_{\pm \mathrm{stdev}}$ over three independent runs.}
\label{tab:main_results}
\small
\setlength{\tabcolsep}{5pt}
\renewcommand{\arraystretch}{1.15}

\begin{adjustbox}{max width=\linewidth}
\begin{tabular}{l*{8}{c}}
\toprule
\textbf{Model}
& \multicolumn{4}{c}{\textbf{P-Easy}}
& \multicolumn{4}{c}{\textbf{P-Hard}} \\
\cmidrule(lr){2-5} \cmidrule(lr){6-9}
& \multicolumn{2}{c}{Raw}
& \multicolumn{2}{c}{Strict}
& \multicolumn{2}{c}{Raw}
& \multicolumn{2}{c}{Strict} \\
\cmidrule(lr){2-3} \cmidrule(lr){4-5}
\cmidrule(lr){6-7} \cmidrule(lr){8-9}
& pass@1 & pass@3
& pass@1 & pass@3
& pass@1 & pass@3
& pass@1 & pass@3 \\
\midrule

\multicolumn{9}{c}{\textit{Proprietary Models}} \\
\midrule
GPT-5.4
& $\mathbf{92.9}_{\pm 1.0}$ & 95.6
& $64.7_{\pm 3.3}$ & 73.4
& $58.3_{\pm 0.9}$ & 68.0
& $30.5_{\pm 1.6}$ & 39.2 \\

\midrule
\multicolumn{9}{c}{\textit{Open-source Models}} \\
\midrule
DeepSeek V4 Pro
& $70.9_{\pm 0.5}$ & 93.1
& $54.2_{\pm 0.5}$ & 71.9
& $46.8_{\pm 1.2}$ & 69.7
& $26.3_{\pm 3.0}$ & 42.5 \\

GPT-OSS-120B
& $48.1_{\pm 4.3}$ & 86.2
& $26.3_{\pm 3.6}$ & 55.2
& $37.4_{\pm 4.3}$ & 71.2
& $18.2_{\pm 3.5}$ & 38.7 \\

Qwen-3-Coder-30B
& $76.9_{\pm 2.2}$ & 87.7
& $45.0_{\pm 1.1}$ & 57.6
& $59.8_{\pm 0.9}$ & 68.0
& $27.8_{\pm 1.8}$ & 36.0 \\

Qwen-3-32B
& $82.8_{\pm 0.0}$ & 90.6
& $46.1_{\pm 1.7}$ & 56.2
& $58.4_{\pm 0.7}$ & 71.6
& $27.2_{\pm 0.7}$ & 38.7 \\

Qwen-2.5-Coder-7B
& $61.4_{\pm 8.5}$ & 83.7
& $36.3_{\pm 5.5}$ & 54.7
& $37.5_{\pm 2.8}$ & 63.1
& $13.2_{\pm 0.5}$ & 25.2 \\
\quad + StatQA-FT
& $70.1_{\pm 1.1}$ & 87.2
& $38.6_{\pm 1.6}$ & 51.2
& $47.1_{\pm 0.3}$ & 65.8
& $21.2_{\pm 2.0}$ & 31.5 \\

\quad +  DARE-retrieval
& $56.7_{\pm 7.3}$ & 83.7
& $32.3_{\pm 5.7}$ & 51.2
& $39.2_{\pm 1.6}$ & 61.7
& $18.5_{\pm 2.0}$ & 32.4 \\
DataMind-7B
& $60.3_{\pm 1.0}$ & 84.2
& $31.5_{\pm 1.0}$ & 49.8
& $44.4_{\pm 4.2}$ & 66.7
& $19.1_{\pm 1.7}$ & 32.0 \\

\rowcolor{orange!14}
\textbf{Fisher-R1-7B}
& $87.0_{\pm 1.6}$ & $\mathbf{96.6}$
& $\mathbf{65.7}_{\pm 1.2}$ & 74.9
& $63.4_{\pm 2.5}$ & $\mathbf{82.4}$
& $30.6_{\pm 1.2}$ & 44.1 \\

Qwen-2.5-Coder-14B
& $78.5_{\pm 1.2}$ & 88.2
& $41.4_{\pm 2.5}$ & 49.8
& $51.5_{\pm 4.3}$ & 66.7
& $23.7_{\pm 1.6}$ & 33.3 \\
\quad + StatQA-FT
& $75.4_{\pm 1.5}$ & 84.2
& $38.9_{\pm 0.5}$ & 46.8
& $55.3_{\pm 1.4}$ & 65.8
& $27.3_{\pm 1.0}$ & 35.1 \\

\quad +  DARE-retrieval
& $84.1_{\pm 1.0}$ & 91.6
& $46.6_{\pm 1.1}$ & 56.7
& $56.9_{\pm 2.3}$ & 70.3
& $29.1_{\pm 1.1}$ & 41.0 \\

DataMind-14B
& $69.0_{\pm 5.2}$ & 88.2
& $39.6_{\pm 3.0}$ & 53.7
& $48.2_{\pm 2.1}$ & 67.1
& $22.1_{\pm 1.2}$ & 33.8 \\

\rowcolor{orange!14}
\textbf{Fisher-R1-14B}
& $87.4_{\pm 1.9}$ & 94.1
& $64.2_{\pm 0.6}$ & $\mathbf{75.9}$
& $\mathbf{65.8}_{\pm 0.8}$ & 81.1
& $\mathbf{33.0}_{\pm 1.7}$ & $\mathbf{45.5}$ \\

\bottomrule
\end{tabular}
\end{adjustbox}
\end{table}

Fisher-R1 turns its Qwen-2.5-Coder backbones into the strongest open-source hypothesis-testing agents on P-Bench (Table~\ref{tab:main_results}). With the 7B backbone, Raw pass@1 rises from 61.4 to 87.0 on P-Easy and from 37.5 to 63.4 on P-Hard, while Strict pass@1 jumps from 36.3 to 65.7 (P-Easy) and 13.2 to 30.6 (P-Hard); the 14B variant achieves the best P-Hard Strict scores in the table (33.0 pass@1, 45.5 pass@3). Run-to-run standard deviations also drop sharply (e.g., $\pm 8.5\to\pm 1.6$ on P-Easy Raw pass@1 for the 7B), so Fisher-R1 is not only more accurate but also more stable across rollouts. At both model scales, Fisher-R1 outperforms StatQA-FT and DARE-retrieval on every metric, showing that neither statistical supervision nor documentation retrieval matches the gains achieved by training with verified statistical rewards.

Fisher-R1 also outperforms substantially larger open-source baselines. Despite using 7B/14B backbones, it beats DeepSeek V4 Pro, GPT-OSS-120B, Qwen-3-Coder-30B, and Qwen-3-32B on every Strict metric, indicating that for hypothesis testing, targeted training with a verified statistical reward is more effective than scaling the underlying coding model. Fisher-R1-14B further exceeds GPT-5.4 on three of four Strict metrics (P-Easy pass@3: 75.9 vs 73.4; P-Hard pass@1: 33.0 vs 30.5; P-Hard pass@3: 45.5 vs 39.2), trailing only narrowly on P-Easy Strict pass@1 (64.2 vs 64.7).

Across all baselines, Strict scores are far below Raw scores, especially on P-Hard. GPT-5.4, for example, scores 58.3 Raw but only 30.5 Strict on P-Hard pass@1: it reports the correct reject/fail-to-reject direction nearly twice as often as it produces a $p$-value close to the canonical analysis. This pattern is robust to the choice of Strict threshold: using a relaxed criterion of $|\Delta z| < 1$ leads to the same qualitative conclusion (Table~\ref{tab:main_results_z1}) as the main criterion of $|\Delta z| < 0.5$. Fisher-R1 shrinks this gap, suggesting that conclusion-only evaluation overstates agent reliability in open-ended hypothesis testing, and that explicitly rewarding $p$-value accuracy is what gets the model to ground its decision in the executed analysis.

\subsection{Ablation Analysis}
Table~\ref{tab:model_reward_components} shows that combining the SFT warm-start with DAPO is essential to reach Fisher-R1's full performance: SFT+DAPO achieves the best score on every metric, substantially above either stage alone. DAPO applied directly to the backbone lifts single-run accuracy (e.g., P-Hard Strict pass@1 13.2 to 25.2), but plateaus well below SFT+DAPO. SFT alone improves pass@3 by broadening the solution distribution, but does not reliably improve pass@1.

\begin{table}[!h]
\centering
\caption{Ablation of SFT warm-start and DAPO training on P-Bench with the 7B backbone.}
\label{tab:model_reward_components}
\small
\setlength{\tabcolsep}{5pt}
\renewcommand{\arraystretch}{1.12}
\begin{adjustbox}{max width=\linewidth}
\begin{tabular}{lcccccccc}
\toprule
\textbf{Backbone}
& \multicolumn{4}{c}{\textbf{P-Easy}}
& \multicolumn{4}{c}{\textbf{P-Hard}} \\
\cmidrule(lr){2-5} \cmidrule(lr){6-9}
& \multicolumn{2}{c}{Raw}
& \multicolumn{2}{c}{Strict}
& \multicolumn{2}{c}{Raw}
& \multicolumn{2}{c}{Strict} \\
\cmidrule(lr){2-3} \cmidrule(lr){4-5}
\cmidrule(lr){6-7} \cmidrule(lr){8-9}
& pass@1 & pass@3
& pass@1 & pass@3
& pass@1 & pass@3
& pass@1 & pass@3 \\
\midrule

\rowcolor{gray!18}
Qwen-2.5-Coder-7B
& $61.4_{\pm 8.5}$ & 83.7
& $36.3_{\pm 5.5}$ & 54.7
& $37.5_{\pm 2.8}$ & 63.1
& $13.2_{\pm 0.5}$ & 25.2 \\
\quad +DAPO                  
  & $81.6_{\pm 2.8}$ & 86.9    
  & $44.4_{\pm 0.9}$ & 48.2 
  & $52.3_{\pm 3.6}$ & 62.6  
  & $25.2_{\pm 0.8}$ & 30.3 \\  

\quad +SFT
 & $51.6_{\pm 6.1}$ & 84.2        
  & $30.2_{\pm 4.6}$ & 59.6
  & $33.5_{\pm 3.9}$ & 68.0
  & $14.3_{\pm 1.1}$ & 30.2 \\      

\rowcolor{orange!14}
\qquad +DAPO
& $87.0_{\pm 1.6}$ & $\mathbf{96.6}$
& $\mathbf{65.7}_{\pm 1.2}$ & 74.9
& $63.4_{\pm 2.5}$ & $\mathbf{82.4}$
& $30.6_{\pm 1.2}$ & 44.1 \\

\bottomrule
\end{tabular}
\end{adjustbox}
\end{table}

\subsection{Failure Analysis}
\label{sec:failure-analysis}

We inspect one fixed rollout from Qwen-2.5-Coder-7B and Fisher-R1-7B on all 425 P-Bench tasks. Table~\ref{tab:failure-comparison} contrasts the two traces. Fisher-R1 corrects 118 backbone failures, with the largest gain coming from the reduction in operational failures. Its remaining errors concentrate in model or design specification (78/213, including 29 incorrect method choices) and perturbation or data quality handling (65/213), while we find no standalone numerical $p$-value error once the analyzed sample, estimand, model specification, weights, and variance estimator are held fixed. Fisher-R1 also improves substantially on Spearman correlation (7/55 to 42/55), log-rank tests (18/35 to 30/35), and Cox regression (5/20 to 12/20), indicating that its gains extend beyond merely completing more trajectories. 

\begin{table}[!h]
\centering
\caption{Trace level comparison of Qwen-2.5-Coder-7B and Fisher-R1-7B on one fixed rollout over all 425 P-Bench tasks. Changes are reported in percentage points (pp).}
\label{tab:failure-comparison}
\small
\renewcommand{\arraystretch}{1.12}
\begin{tabular*}{\linewidth}{@{\extracolsep{\fill}}lrrr@{}}
\toprule
\textbf{Outcome} & \textbf{Qwen-2.5-Coder-7B} & \textbf{Fisher-R1-7B} & \textbf{Change} \\
\midrule
Strict success                 & 124 (29.2\%) & \textbf{212 (49.9\%)} & $+20.7$ pp \\
Operational failure            & 185 (43.5\%) & \textbf{36 (8.5\%)}   & $-35.0$ pp \\
Method selection error         &  50 (11.8\%) & \textbf{29 (6.8\%)}   & $-5.0$ pp  \\
Clean task Strict accuracy     & 33.8\%        & \textbf{57.2\%}        & $+23.4$ pp \\
Perturbed task Strict accuracy & 12.1\%        & \textbf{23.1\%}        & $+11.0$ pp \\
\bottomrule
\end{tabular*}
\end{table}

\subsection{Generalization, not Memorization}

To verify that gains on P-Bench reflect generalization rather than memorization, we quantify the distance between the evaluation set and the SFT and RL training corpus in semantic-embedding space. For each input, we extract the research question and study description, i.e., the user prompt, and embed it with OpenAI's \texttt{text-embedding-3-small} \citep{openai2026textembedding3small}. We then compute, for each sample, the mean cosine similarity to its top-5 nearest neighbors in the training pool.

\begin{wrapfigure}{r}{0.5\textwidth}
    \centering
    \vspace{-0.5em}
    \includegraphics[width=0.5\textwidth]{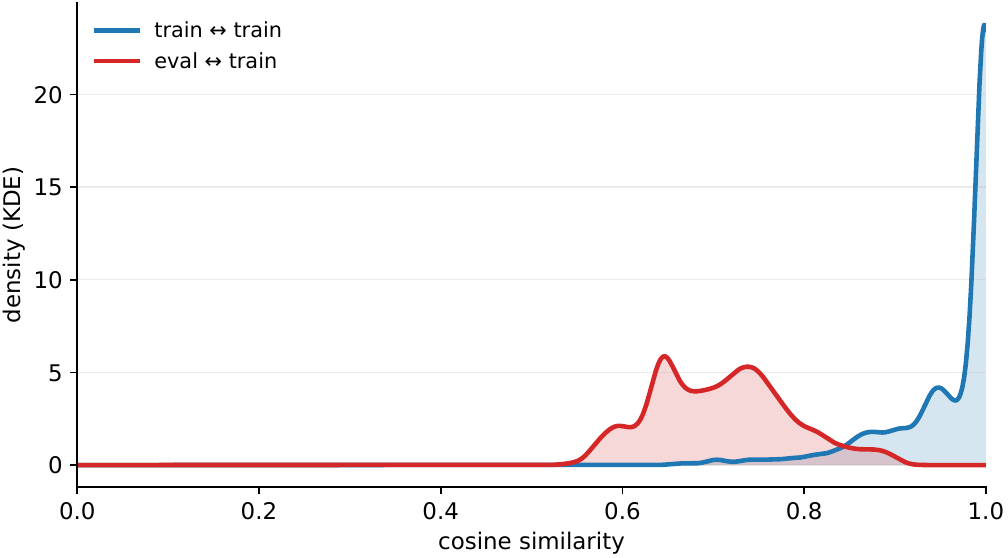}
    \caption{Similarity Analysis}
    \label{fig:data_gap}
    \vspace{-1em}
\end{wrapfigure}
As shown in Fig.~\ref{fig:data_gap}, train-to-train similarities form a tight high-similarity ``memorization band'' that reflects the templated structure of generated RL prompts, while eval-to-train similarities sit clearly below this band. The 425 P-Bench tasks, therefore, have no near-duplicate counterparts in the RL corpus, indicating that gains on P-Bench reflect generalization rather than retrieval of training prompts. The two distributions also differ by construction: training tasks are synthetic simulations, whereas P-Bench tasks are real-world hypothesis-testing tasks.

\section{Discussion}
We introduced P-Bench, a benchmark of 425 real-world hypothesis-testing tasks with expert-audited answer keys, and Fisher-R1, an open-weight agent trained on synthetic statistical tasks via SFT and RL with verified-outcome rewards. Our results show that even current frontier models still lack reliable statistical reasoning for open-ended hypothesis testing. Fisher-R1-14B outperforms strong proprietary and open-source baselines, including GPT-5.4 and DeepSeek-V4-Pro. Our evaluation is limited to frequentist, single-hypothesis decisions at a prespecified significance level of $\alpha=0.05$, and does not assess threshold selection, multiple-comparison correction, Bayesian evidence, or the practical significance of effect estimates; extending verified rewards to these settings remains important future work. Reliable statistical agents can improve scientific reproducibility by catching method-driven errors before they enter the literature. They could also lend false legitimacy to weak claims. We therefore release P-Bench and Fisher-R1 as evaluation and oversight tools, not as substitutes for human statistical review. Future work should further study how to make agents explicitly justify assumptions, quantify uncertainty over method choice, and recognize when no single hypothesis test is adequate for a scientific question.

\newpage
\bibliographystyle{unsrtnat}
\bibliography{ref}

@article{Wasserstein02042016,
author = {Ronald L. Wasserstein and Nicole A. Lazar},
title = {The ASA Statement on p-Values: Context, Process, and Purpose},
journal = {The American Statistician},
volume = {70},
number = {2},
pages = {129--133},
year = {2016},
publisher = {Taylor \& Francis},
doi = {10.1080/00031305.2016.1154108},
URL = { 
    
        https://doi.org/10.1080/00031305.2016.1154108

},
eprint = { 
    
        https://doi.org/10.1080/00031305.2016.1154108

}
}

@article{zhu2024large,
  title={Are large language models good statisticians?},
  author={Zhu, Yizhang and Du, Shiyin and Li, Boyan and Luo, Yuyu and Tang, Nan},
  journal={Advances in Neural Information Processing Systems},
  volume={37},
  pages={62697--62731},
  year={2024}
}

@inproceedings{hong2025data,
  title={Data interpreter: An llm agent for data science},
  author={Hong, Sirui and Lin, Yizhang and Liu, Bang and Liu, Bangbang and Wu, Binhao and Zhang, Ceyao and Li, Danyang and Chen, Jiaqi and Zhang, Jiayi and Wang, Jinlin and others},
  booktitle={Findings of the Association for Computational Linguistics: ACL 2025},
  pages={19796--19821},
  year={2025}
}

@article{zhang2025deepanalyze,
  title={Deepanalyze: Agentic large language models for autonomous data science},
  author={Zhang, Shaolei and Fan, Ju and Fan, Meihao and Li, Guoliang and Du, Xiaoyong},
  journal={arXiv preprint arXiv:2510.16872},
  year={2025}
}

@article{li2024tapilot,
  title={Tapilot-crossing: Benchmarking and evolving llms towards interactive data analysis agents},
  author={Li, Jinyang and Huo, Nan and Gao, Yan and Shi, Jiayi and Zhao, Yingxiu and Qu, Ge and Wu, Yurong and Ma, Chenhao and Lou, Jian-Guang and Cheng, Reynold},
  journal={arXiv preprint arXiv:2403.05307},
  year={2024}
}

@article{knaus1995support,
  title={The SUPPORT prognostic model: Objective estimates of survival for seriously ill hospitalized adults},
  author={Knaus, William A and Harrell, Frank E and Lynn, Joanne and Goldman, Lee and Phillips, Russell S and Connors, Alfred F and Dawson, Neal V and Fulkerson, William J and Califf, Robert M and Desbiens, Norman and others},
  journal={Annals of internal medicine},
  volume={122},
  number={3},
  pages={191--203},
  year={1995},
  publisher={American College of Physicians}
}

@article{mckelway2023effects,
  title={Effects of cognitive behavioral therapy and cash transfers on older persons living alone in India: a randomized trial},
  author={McKelway, Madeline and Banerjee, Abhijit and Grela, Erin and Schilbach, Frank and Sequeira, Miriam and Sharma, Garima and Vaidyanathan, Girija and Duflo, Esther},
  journal={Annals of internal medicine},
  volume={176},
  number={5},
  pages={632--641},
  year={2023},
  publisher={American College of Physicians}
}

@article{nguyen2022genomic,
  title={Genomic characterization of metastatic patterns from prospective clinical sequencing of 25,000 patients},
  author={Nguyen, Bastien and Fong, Christopher and Luthra, Anisha and Smith, Shaleigh A and DiNatale, Renzo G and Nandakumar, Subhiksha and Walch, Henry and Chatila, Walid K and Madupuri, Ramyasree and Kundra, Ritika and others},
  journal={Cell},
  volume={185},
  number={3},
  pages={563--575},
  year={2022},
  publisher={Elsevier}
}

@article{yamada2025ai,
  title={The ai scientist-v2: Workshop-level automated scientific discovery via agentic tree search},
  author={Yamada, Yutaro and Lange, Robert Tjarko and Lu, Cong and Hu, Shengran and Lu, Chris and Foerster, Jakob and Clune, Jeff and Ha, David},
  journal={arXiv preprint arXiv:2504.08066},
  year={2025}
}

@article{tang2025ai,
  title={Ai-researcher: Autonomous scientific innovation},
  author={Tang, Jiabin and Xia, Lianghao and Li, Zhonghang and Huang, Chao},
  journal={arXiv preprint arXiv:2505.18705},
  year={2025}
}

@article{lu2026towards,
  title={Towards end-to-end automation of AI research},
  author={Lu, Chris and Lu, Cong and Lange, Robert Tjarko and Yamada, Yutaro and Hu, Shengran and Foerster, Jakob and Ha, David and Clune, Jeff},
  journal={Nature},
  volume={651},
  number={8107},
  pages={914--919},
  year={2026},
  publisher={Nature Publishing Group UK London}
}

@inproceedings{lal2025musciclaims,
  title={MuSciClaims: Multimodal Scientific Claim Verification},
  author={Lal, Yash Kumar and Bandham, Manikanta and Hasan, Mohammad Saqib and Kashi, Apoorva and Koupaee, Mahnaz and Balasubramanian, Niranjan},
  booktitle={Proceedings of the 14th International Joint Conference on Natural Language Processing and the 4th Conference of the Asia-Pacific Chapter of the Association for Computational Linguistics},
  pages={3285--3307},
  year={2025}
}

@inproceedings{kumar2025sciclaimhunt,
  title={Sciclaimhunt: A large dataset for evidence-based scientific claim verification},
  author={Kumar, Sujit and Sharma, Anshul and Khincha, Siddharth Hemant and Shroff, Gargi and Singh, Sanasam Ranbir and Mishra, Rahul},
  booktitle={2025 International Joint Conference on Neural Networks (IJCNN)},
  pages={1--10},
  year={2025},
  organization={IEEE}
}

@article{li2025ida,
  title={IDA-Bench: Evaluating LLMs on Interactive Guided Data Analysis},
  author={Li, Hanyu and Liu, Haoyu and Zhu, Tingyu and Guo, Tianyu and Zheng, Zeyu and Deng, Xiaotie and Jordan, Michael I},
  journal={arXiv preprint arXiv:2505.18223},
  year={2025}
}

@misc{dfeep_harvard_dataverse,
  author       = {{Abdul Latif Jameel Poverty Action Lab (J-PAL)}},
  title        = {Development Economics Field Experiments (DFEEP)},
  howpublished = {Harvard Dataverse},
  year         = {2026},
  note         = {Accessed: 2026-04-20},
  url          = {https://dataverse.harvard.edu/dataverse/DFEEP}
}

@article{cerami2012cbio,
  title={The cBio cancer genomics portal: an open platform for exploring multidimensional cancer genomics data},
  author={Cerami, Ethan and Gao, Jianjiong and Dogrusoz, Ugur and Gross, Benjamin E and Sumer, Selcuk Onur and Aksoy, B{\"u}lent Arman and Jacobsen, Anders and Byrne, Caitlin J and Heuer, Michael L and Larsson, Erik and others},
  journal={Cancer discovery},
  volume={2},
  number={5},
  pages={401--404},
  year={2012},
  publisher={American Association for Cancer Research}
}

@article{gao2013integrative,
  title={Integrative analysis of complex cancer genomics and clinical profiles using the cBioPortal},
  author={Gao, Jianjiong and Aksoy, B{\"u}lent Arman and Dogrusoz, Ugur and Dresdner, Gideon and Gross, Benjamin and Sumer, S Onur and Sun, Yichao and Jacobsen, Anders and Sinha, Rileen and Larsson, Erik and others},
  journal={Science signaling},
  volume={6},
  number={269},
  pages={pl1--pl1},
  year={2013},
  publisher={American Association for the Advancement of Science}
}

@misc{wang2025biodsa1kbenchmarkingdatascience,
      title={BioDSA-1K: Benchmarking Data Science Agents for Biomedical Research}, 
      author={Zifeng Wang and Benjamin Danek and Jimeng Sun},
      year={2025},
      eprint={2505.16100},
      archivePrefix={arXiv},
      primaryClass={cs.AI},
      url={https://arxiv.org/abs/2505.16100}, 
}

@misc{harrell_hbiostat,
    title        = {Vanderbilt Biostatistics Datasets (hbiostat.org)},
    author       = {Harrell, Frank E.},
    howpublished = {\url{https://hbiostat.org/data/}},
    year         = {2024},
    note         = {Accessed 2026-04-20}
  }

@article{yu2025dapo,
  title={Dapo: An open-source llm reinforcement learning system at scale},
  author={Yu, Qiying and Zhang, Zheng and Zhu, Ruofei and Yuan, Yufeng and Zuo, Xiaochen and Yue, Yu and Dai, Weinan and Fan, Tiantian and Liu, Gaohong and Liu, Lingjun and others},
  journal={arXiv preprint arXiv:2503.14476},
  year={2025}
}

@inproceedings{lai2023ds,
  title={DS-1000: A natural and reliable benchmark for data science code generation},
  author={Lai, Yuhang and Li, Chengxi and Wang, Yiming and Zhang, Tianyi and Zhong, Ruiqi and Zettlemoyer, Luke and Yih, Wen-tau and Fried, Daniel and Wang, Sida and Yu, Tao},
  booktitle={International Conference on Machine Learning},
  pages={18319--18345},
  year={2023},
  organization={PMLR}
}

@article{hu2024infiagent,
  title={Infiagent-dabench: Evaluating agents on data analysis tasks},
  author={Hu, Xueyu and Zhao, Ziyu and Wei, Shuang and Chai, Ziwei and Ma, Qianli and Wang, Guoyin and Wang, Xuwu and Su, Jing and Xu, Jingjing and Zhu, Ming and others},
  journal={arXiv preprint arXiv:2401.05507},
  year={2024}
}

@article{zhang2025datascibench,
  title={Datascibench: An llm agent benchmark for data science},
  author={Zhang, Dan and Zhoubian, Sining and Cai, Min and Li, Fengzu and Yang, Lekang and Wang, Wei and Dong, Tianjiao and Hu, Ziniu and Tang, Jie and Yue, Yisong},
  journal={arXiv preprint arXiv:2502.13897},
  year={2025}
}

@article{chan2024mle,
  title={Mle-bench: Evaluating machine learning agents on machine learning engineering},
  author={Chan, Jun Shern and Chowdhury, Neil and Jaffe, Oliver and Aung, James and Sherburn, Dane and Mays, Evan and Starace, Giulio and Liu, Kevin and Maksin, Leon and Patwardhan, Tejal and others},
  journal={arXiv preprint arXiv:2410.07095},
  year={2024}
}

@inproceedings{wadden2020fact,
  title={Fact or fiction: Verifying scientific claims},
  author={Wadden, David and Lin, Shanchuan and Lo, Kyle and Wang, Lucy Lu and van Zuylen, Madeleine and Cohan, Arman and Hajishirzi, Hannaneh},
  booktitle={Proceedings of the 2020 Conference on Empirical Methods in Natural Language Processing (EMNLP)},
  pages={7534--7550},
  year={2020}
}

@article{miske2026investigating,
  title={Investigating the reproducibility of the social and behavioural sciences},
  author={Miske, Olivia and Abatayo, Anna Lou and Daley, Mason and Dirzo, Mirka and Fox, Nicholas and Haber, Noah and Hahn, Krystal M and Struhl, Melissa Kline and Mawhinney, Brinna and Silverstein, Priya and others},
  journal={Nature},
  volume={652},
  number={8108},
  pages={126--134},
  year={2026},
  publisher={Nature Publishing Group UK London}
}

@misc{hu2024infiagentdabenchevaluatingagentsdata,
      title={InfiAgent-DABench: Evaluating Agents on Data Analysis Tasks}, 
      author={Xueyu Hu and Ziyu Zhao and Shuang Wei and Ziwei Chai and Qianli Ma and Guoyin Wang and Xuwu Wang and Jing Su and Jingjing Xu and Ming Zhu and Yao Cheng and Jianbo Yuan and Jiwei Li and Kun Kuang and Yang Yang and Hongxia Yang and Fei Wu},
      year={2024},
      eprint={2401.05507},
      archivePrefix={arXiv},
      primaryClass={cs.CL},
      url={https://arxiv.org/abs/2401.05507}, 
}

@article{egg2025dabstep,
  title={Dabstep: Data agent benchmark for multi-step reasoning},
  author={Egg, Alex and Goyanes, Martin Iglesias and Kingma, Friso and Mora, Andreu and von Werra, Leandro and Wolf, Thomas},
  journal={arXiv preprint arXiv:2506.23719},
  year={2025}
}

@inproceedings{huang-etal-2024-da,
    title = "{DA}-Code: Agent Data Science Code Generation Benchmark for Large Language Models",
    author = "Huang, Yiming  and
      Luo, Jianwen  and
      Yu, Yan  and
      Zhang, Yitong  and
      Lei, Fangyu  and
      Wei, Yifan  and
      He, Shizhu  and
      Huang, Lifu  and
      Liu, Xiao  and
      Zhao, Jun  and
      Liu, Kang",
    editor = "Al-Onaizan, Yaser  and
      Bansal, Mohit  and
      Chen, Yun-Nung",
    booktitle = "Proceedings of the 2024 Conference on Empirical Methods in Natural Language Processing",
    month = nov,
    year = "2024",
    address = "Miami, Florida, USA",
    publisher = "Association for Computational Linguistics",
    url = "https://aclanthology.org/2024.emnlp-main.748/",
    doi = "10.18653/v1/2024.emnlp-main.748",
    pages = "13487--13521"
}

@article{zhang2023data,
  title={Data-copilot: Bridging billions of data and humans with autonomous workflow},
  author={Zhang, Wenqi and Shen, Yongliang and Lu, Weiming and Zhuang, Yueting},
  journal={arXiv preprint arXiv:2306.07209},
  year={2023}
}

@misc{hui2024qwen25codertechnicalreport,
      title={Qwen2.5-Coder Technical Report}, 
      author={Binyuan Hui and Jian Yang and Zeyu Cui and Jiaxi Yang and Dayiheng Liu and Lei Zhang and Tianyu Liu and Jiajun Zhang and Bowen Yu and Keming Lu and Kai Dang and Yang Fan and Yichang Zhang and An Yang and Rui Men and Fei Huang and Bo Zheng and Yibo Miao and Shanghaoran Quan and Yunlong Feng and Xingzhang Ren and Xuancheng Ren and Jingren Zhou and Junyang Lin},
      year={2024},
      eprint={2409.12186},
      archivePrefix={arXiv},
      primaryClass={cs.CL},
      url={https://arxiv.org/abs/2409.12186}, 
}

@misc{openai2026gpt54,
  author       = {OpenAI},
  title        = {Introducing GPT-5.4},
  year         = {2026},
  month        = mar,
  day          = {5},
  howpublished = {\url{https://openai.com/index/introducing-gpt-5-4/}},
  note         = {Accessed: 2026-05-06}
}

@misc{deepseek2026v4,
  author       = {{DeepSeek-AI}},
  title        = {DeepSeek-V4: Towards Highly Efficient Million-Token Context Intelligence},
  year         = {2026},
  month        = apr,
  howpublished = {\url{https://huggingface.co/deepseek-ai/DeepSeek-V4-Pro}},
  note         = {Model card and technical report for DeepSeek-V4-Pro. Accessed: 2026-05-06}
}

@article{qwen3,
  title   = {Qwen3 Technical Report},
  author  = {An Yang and Anfeng Li and Baosong Yang and Beichen Zhang and Binyuan Hui and Bo Zheng and Bowen Yu and Chang Gao and Chengen Huang and Chenxu Lv and Chujie Zheng and Dayiheng Liu and Fan Zhou and Fei Huang and Feng Hu and Hao Ge and Haoran Wei and Huan Lin and Jialong Tang and Jian Yang and Jianhong Tu and Jianwei Zhang and Jianxin Yang and Jiaxi Yang and Jing Zhou and Jingren Zhou and Junyang Lin and Kai Dang and Keqin Bao and Kexin Yang and Le Yu and Lianghao Deng and Mei Li and Mingfeng Xue and Mingze Li and Pei Zhang and Peng Wang and Qin Zhu and Rui Men and Ruize Gao and Shixuan Liu and Shuang Luo and Tianhao Li and Tianyi Tang and Wenbiao Yin and Xingzhang Ren and Xinyu Wang and Xinyu Zhang and Xuancheng Ren and Yang Fan and Yang Su and Yichang Zhang and Yinger Zhang and Yu Wan and Yuqiong Liu and Zekun Wang and Zeyu Cui and Zhenru Zhang and Zhipeng Zhou and Zihan Qiu},
  journal = {arXiv preprint arXiv:2505.09388},
  year    = {2025}
}

@misc{qwen3coder30b,
  author       = {{Qwen Team}},
  title        = {Qwen3-Coder-30B-A3B-Instruct},
  year         = {2025},
  howpublished = {\url{https://huggingface.co/Qwen/Qwen3-Coder-30B-A3B-Instruct}},
  note         = {Model card. Accessed: 2026-05-06}
}

@misc{qiao2026scalinggeneralistdataanalyticagents,
      title={Scaling Generalist Data-Analytic Agents}, 
      author={Shuofei Qiao and Yanqiu Zhao and Zhisong Qiu and Xiaobin Wang and Jintian Zhang and Zhao Bin and Ningyu Zhang and Yong Jiang and Pengjun Xie and Fei Huang and Huajun Chen},
      year={2026},
      eprint={2509.25084},
      archivePrefix={arXiv},
      primaryClass={cs.CL},
      url={https://arxiv.org/abs/2509.25084}, 
}

@misc{zheng2024llamafactoryunifiedefficientfinetuning,
      title={LlamaFactory: Unified Efficient Fine-Tuning of 100+ Language Models}, 
      author={Yaowei Zheng and Richong Zhang and Junhao Zhang and Yanhan Ye and Zheyan Luo and Zhangchi Feng and Yongqiang Ma},
      year={2024},
      eprint={2403.13372},
      archivePrefix={arXiv},
      primaryClass={cs.CL},
      url={https://arxiv.org/abs/2403.13372}, 
}

@inproceedings{sheng2025hybridflow,
  title={Hybridflow: A flexible and efficient rlhf framework},
  author={Sheng, Guangming and Zhang, Chi and Ye, Zilingfeng and Wu, Xibin and Zhang, Wang and Zhang, Ru and Peng, Yanghua and Lin, Haibin and Wu, Chuan},
  booktitle={Proceedings of the Twentieth European Conference on Computer Systems},
  pages={1279--1297},
  year={2025}
}

@misc{madaan2023selfrefineiterativerefinementselffeedback,
      title={Self-Refine: Iterative Refinement with Self-Feedback}, 
      author={Aman Madaan and Niket Tandon and Prakhar Gupta and Skyler Hallinan and Luyu Gao and Sarah Wiegreffe and Uri Alon and Nouha Dziri and Shrimai Prabhumoye and Yiming Yang and Shashank Gupta and Bodhisattwa Prasad Majumder and Katherine Hermann and Sean Welleck and Amir Yazdanbakhsh and Peter Clark},
      year={2023},
      eprint={2303.17651},
      archivePrefix={arXiv},
      primaryClass={cs.CL},
      url={https://arxiv.org/abs/2303.17651}, 
}

@article{shinn2023reflexion,
  title={Reflexion: Language agents with verbal reinforcement learning},
  author={Shinn, Noah and Cassano, Federico and Gopinath, Ashwin and Narasimhan, Karthik and Yao, Shunyu},
  journal={Advances in neural information processing systems},
  volume={36},
  pages={8634--8652},
  year={2023}
}

@article{wang2023voyager,
  title={Voyager: An open-ended embodied agent with large language models},
  author={Wang, Guanzhi and Xie, Yuqi and Jiang, Yunfan and Mandlekar, Ajay and Xiao, Chaowei and Zhu, Yuke and Fan, Linxi and Anandkumar, Anima},
  journal={arXiv preprint arXiv:2305.16291},
  year={2023}
}

@article{yang2024swe,
  title={Swe-agent: Agent-computer interfaces enable automated software engineering},
  author={Yang, John and Jimenez, Carlos E and Wettig, Alexander and Lieret, Kilian and Yao, Shunyu and Narasimhan, Karthik and Press, Ofir},
  journal={Advances in Neural Information Processing Systems},
  volume={37},
  pages={50528--50652},
  year={2024}
}

@article{schulman2017proximal,
  title={Proximal policy optimization algorithms},
  author={Schulman, John and Wolski, Filip and Dhariwal, Prafulla and Radford, Alec and Klimov, Oleg},
  journal={arXiv preprint arXiv:1707.06347},
  year={2017}
}

@article{ouyang2022training,
  title={Training language models to follow instructions with human feedback},
  author={Ouyang, Long and Wu, Jeffrey and Jiang, Xu and Almeida, Diogo and Wainwright, Carroll and Mishkin, Pamela and Zhang, Chong and Agarwal, Sandhini and Slama, Katarina and Ray, Alex and others},
  journal={Advances in neural information processing systems},
  volume={35},
  pages={27730--27744},
  year={2022}
}

@article{rafailov2023direct,
  title={Direct preference optimization: Your language model is secretly a reward model},
  author={Rafailov, Rafael and Sharma, Archit and Mitchell, Eric and Manning, Christopher D and Ermon, Stefano and Finn, Chelsea},
  journal={Advances in neural information processing systems},
  volume={36},
  pages={53728--53741},
  year={2023}
}

@article{shao2024deepseekmath,
  title={Deepseekmath: Pushing the limits of mathematical reasoning in open language models},
  author={Shao, Zhihong and Wang, Peiyi and Zhu, Qihao and Xu, Runxin and Song, Junxiao and Bi, Xiao and Zhang, Haowei and Zhang, Mingchuan and Li, YK and Wu, Yang and others},
  journal={arXiv preprint arXiv:2402.03300},
  year={2024}
}

@article{guo2025deepseek,
  title={Deepseek-r1: Incentivizing reasoning capability in llms via reinforcement learning},
  author={Guo, Daya and Yang, Dejian and Zhang, Haowei and Song, Junxiao and Wang, Peiyi and Zhu, Qihao and Xu, Runxin and Zhang, Ruoyu and Ma, Shirong and Bi, Xiao and others},
  journal={arXiv preprint arXiv:2501.12948},
  year={2025}
}

@article{li2025torl,
  title={Torl: Scaling tool-integrated rl},
  author={Li, Xuefeng and Zou, Haoyang and Liu, Pengfei},
  journal={arXiv preprint arXiv:2503.23383},
  year={2025}
}

@article{feng2025retool,
  title={Retool: Reinforcement learning for strategic tool use in llms},
  author={Feng, Jiazhan and Huang, Shijue and Qu, Xingwei and Zhang, Ge and Qin, Yujia and Zhong, Baoquan and Jiang, Chengquan and Chi, Jinxin and Zhong, Wanjun},
  journal={arXiv preprint arXiv:2504.11536},
  year={2025}
}

@article{qian2025toolrl,
  title={Toolrl: Reward is all tool learning needs},
  author={Qian, Cheng and Acikgoz, Emre Can and He, Qi and Wang, Hongru and Chen, Xiusi and Hakkani-T{\"u}r, Dilek and Tur, Gokhan and Ji, Heng},
  journal={arXiv preprint arXiv:2504.13958},
  year={2025}
}

@article{da2025agent,
  title={Agent-rlvr: Training software engineering agents via guidance and environment rewards},
  author={Da, Jeff and Wang, Clinton and Deng, Xiang and Ma, Yuntao and Barhate, Nikhil and Hendryx, Sean},
  journal={arXiv preprint arXiv:2506.11425},
  year={2025}
}

@misc{openai2026textembedding3small,
  author       = {{OpenAI}},
  title        = {\texttt{text-embedding-3-small} Model},
  year         = {2026},
  howpublished = {\url{https://developers.openai.com/api/docs/models/text-embedding-3-small}},
  note         = {Accessed: 2026-05-06}
}

@misc{anthropic2026claudesonnet46,
  title        = {Introducing Claude Sonnet 4.6},
  author       = {{Anthropic}},
  year         = {2026},
  month        = feb,
  howpublished = {\url{https://www.anthropic.com/news/claude-sonnet-4-6}},
  note         = {Accessed: 2026-05-06}
}

@article{yao2022react,
  title={React: Synergizing reasoning and acting in language models},
  author={Yao, Shunyu and Zhao, Jeffrey and Yu, Dian and Du, Nan and Shafran, Izhak and Narasimhan, Karthik and Cao, Yuan},
  journal={arXiv preprint arXiv:2210.03629},
  year={2022}
}

@inproceedings{wang2024executable,
  title={Executable code actions elicit better llm agents},
  author={Wang, Xingyao and Chen, Yangyi and Yuan, Lifan and Zhang, Yizhe and Li, Yunzhu and Peng, Hao and Ji, Heng},
  booktitle={Forty-first International Conference on Machine Learning},
  year={2024}
}

@misc{openai2025gptoss,
  title        = {Introducing gpt-oss},
  author       = {{OpenAI}},
  year         = {2025},
  howpublished = {\url{https://openai.com/index/introducing-gpt-oss/}},
  note         = {Accessed: 2026-05-07}
}

@misc{sun2026darealigningllmagents,
      title={DARE: Aligning LLM Agents with the R Statistical Ecosystem via Distribution-Aware Retrieval}, 
      author={Maojun Sun and Yue Wu and Yifei Xie and Ruijian Han and Binyan Jiang and Defeng Sun and Yancheng Yuan and Jian Huang},
      year={2026},
      eprint={2603.04743},
      archivePrefix={arXiv},
      primaryClass={cs.IR},
      url={https://arxiv.org/abs/2603.04743}, 
}

\newpage
\appendix

\section{Additional Related Works}
\paragraph{Agentic execution and reinforcement learning.}
Beyond static prompting, recent LLM agents solve tasks through iterative interaction with external environments, using tool observations to revise plans, correct errors, and sustain long-horizon execution \citep{madaan2023selfrefineiterativerefinementselffeedback,shinn2023reflexion,wang2023voyager,yang2024swe}. Reinforcement learning has become a central post-training approach for improving such reasoning and tool-use behaviors, evolving from preference-based alignment \citep{schulman2017proximal,ouyang2022training,rafailov2023direct} to outcome-verifiable reasoning: DeepSeekMath introduces GRPO for critic-free RL, DeepSeek-R1 scales RL to induce extended reasoning, and DAPO improves large-scale RL training with decoupled clipping and dynamic sampling \citep{shao2024deepseekmath,guo2025deepseek,yu2025dapo}. A related line studies RL for tool-using agents that decide when and how to call external tools during multi-turn problem solving \citep{li2025torl,feng2025retool,qian2025toolrl,da2025agent}. Whereas these systems reward correctness of a final answer in math, coding, or SQL, Fisher-R1's reward must score \emph{inferential validity}: a p-value compared on a $z$-score scale, plus consistency between the chosen method, the computed evidence, and the reported reject/fail-to-reject decision.

\section{More details on P-bench}
\subsection{Data curation pipeline.}
\begin{wrapfigure}{r}{0.5\textwidth}
    \centering
    \vspace{-0.5em}
    \includegraphics[width=0.5\textwidth]{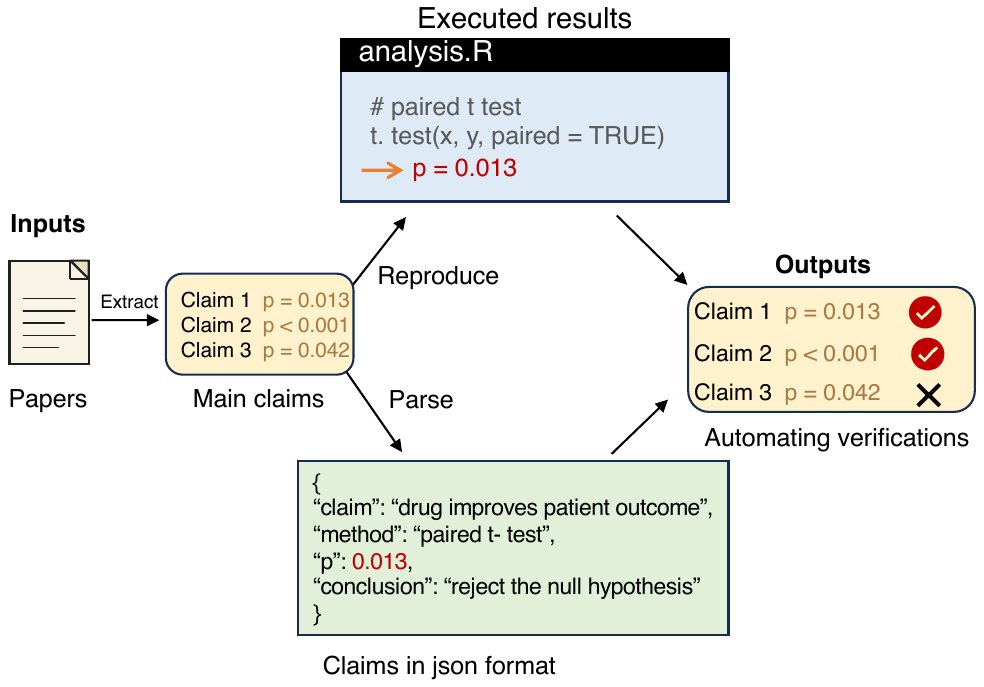}
    \caption{Data curation for P-bench}
    \label{fig:autoclaims_pbench_details}
    \vspace{-1em}
\end{wrapfigure}
Figure~\ref{fig:autoclaims_pbench_details} summarizes how we convert scientific artifacts into verified hypothesis-testing tasks. Given a paper, released data, and available analysis code. It identifies statistical claims with inferential results and parses them into structured records, and links each claim to the corresponding executable analysis. Our system is built on Claude Code with Sonnet 4.5.

\paragraph{Verification and filtering.}
The agent reproduces the analysis in a clean execution environment and extracts p-values,
test statistics, and conclusions from the execution logs rather than from paper text alone. Claims are kept only when the reported result can be matched to an executed computation with a consistent hypothesis-testing decision; ambiguous, failed, or inconsistent cases are discarded. The remaining
claims are further checked by expert review and packaged as P-Bench tasks with hidden answer keys.

\subsection{P-bench Examples}

We provide one representative task from each of the three source families. Each example shows the analysis request and the hidden answer key; the dataset itself is provided to the agent as a CSV file.

\subsection{Example from economics}

\textit{Source: McKelway, Madeline, et al. "Effects of cognitive behavioral therapy and cash transfers on older persons living alone in India: a randomized trial." Annals of internal medicine 176.5 (2023): 632-641. \citep{mckelway2023effects}} A randomized controlled trial assigning elderly persons living alone to one of four arms (control, CBT only, cash only, both) with three measurement waves.

\begin{tcolorbox}[colback=lightboxgray, colframe=boxgray, boxrule=0.6pt, arc=3pt,
  left=8pt, right=8pt, top=4pt, bottom=4pt,
  title=\texttt{analysis\_request.txt}, fonttitle=\bfseries\small, coltitle=white, colbacktitle=boxgray]
\small
\textbf{Research question:} Does phone-based cognitive behavioral therapy without cash support reduce depression among elderly persons living alone in Tamil Nadu, India at 3 weeks after the intervention, relative to the control group?\\[3pt]
\textbf{Hypothesis:}\\
$H_0$: CBT does not affect depression at 3 weeks relative to the control group.\\
$H_1$: CBT only reduces depression at 3 weeks relative to the control group.\\[3pt]
\textbf{Study design:} Randomized controlled trial; individual-level randomization; three waves (baseline, 3 weeks, 3 months).\\[3pt]
\textbf{Variable glossary:} \texttt{GDS} (Geriatric Depression Scale, integer 0--15) [OUTCOME]; \texttt{itreat} (treatment arm, 0--3) [TREATMENT]; \texttt{T} (time period, 0--2); \texttt{elderly\_ID\_anon} (individual id); plus covariates.
\end{tcolorbox}

\begin{tcolorbox}[
  colback=lightboxgray,
  colframe=boxgray,
  boxrule=0.6pt,
  arc=3pt,
  left=8pt,
  right=8pt,
  top=4pt,
  bottom=4pt,
  title={\ttfamily dataset.csv\ (first 3 rows, selected columns)},
  fonttitle=\bfseries\small,
  coltitle=white,
  colbacktitle=boxgray
]

\ttfamily\scriptsize

elderly\_ID\_anon, T, itreat, GDS, ADL, bl\_female\\
211101, 2, 0, 8, 22, 1\\
143101, 2, 0, 8, 4, 1\\
232501, 2, 0, 9, 14, 1\\
\dots

\end{tcolorbox}

\begin{tcolorbox}[colback=lightboxgray, colframe=boxgray, boxrule=0.6pt, arc=3pt,
  left=8pt, right=8pt, top=4pt, bottom=4pt,
  title=\texttt{answer\_key.json} (hidden), fonttitle=\bfseries\small, coltitle=white, colbacktitle=boxgray]
\ttfamily\scriptsize
\{\\
\quad "test\_type": "mixed-effects linear regression",\\
\quad "treatment\_variable": "1.T\#1.itreat (CBT only x 3 weeks)",\\
\quad "treatment\_coefficient": -0.0873,\\
\quad "treatment\_p\_value": 0.876,\\
\quad "significant\_at\_005": false,\\
\quad "conclusion": "Fail to reject H0",\\
\quad "n\_observations": 3009,\\
\quad "n\_groups": 1113\\
\}
\end{tcolorbox}

\subsection{Example from biology}

\textit{Source: Nguyen et al., ``Genomic characterization of metastatic patterns from prospective clinical sequencing of 25{,}000 patients.'' Cell, 2022 \citep{nguyen2022genomic}}.

\begin{tcolorbox}[colback=lightboxgray, colframe=boxgray, boxrule=0.6pt, arc=3pt,
  left=8pt, right=8pt, top=4pt, bottom=4pt,
  title=\texttt{analysis\_request.txt}, fonttitle=\bfseries\small, coltitle=white, colbacktitle=boxgray]
\small
\textbf{Research question:} Is PTEN mutation status associated with metastatic disease among colorectal cancer patients in this tumor sequencing cohort, accounting for basic patient demographics?\\[3pt]
\textbf{Hypothesis:}\\
$H_0$: PTEN mutation status is not associated with metastatic disease among colorectal cancer patients.\\
$H_1$: PTEN mutation status is associated with metastatic disease among colorectal cancer patients.\\[3pt]
\textbf{Study design:} Observational tumor sequencing cohort; 2{,}396 primary tumor samples; metastatic status recorded at the patient level.\\[3pt]
\textbf{Variable glossary:} \texttt{is\_metastatic\_patient} (binary) [OUTCOME]; \texttt{pten\_mutated} (binary) [TARGET]; \texttt{sex\_numeric}, \texttt{age\_at\_sequencing} (covariates).
\end{tcolorbox}

\begin{tcolorbox}[
  colback=lightboxgray,
  colframe=boxgray,
  boxrule=0.6pt,
  arc=3pt,
  left=8pt,
  right=8pt,
  top=4pt,
  bottom=4pt,
  title={\ttfamily dataset.csv\ (first 3 rows, selected columns)},
  fonttitle=\bfseries\small,
  coltitle=white,
  colbacktitle=boxgray
]

\ttfamily\scriptsize

sample\_id, is\_metastatic\_patient, pten\_mutated, sex\_numeric, age\_at\_sequencing\\
P-0000520-T01-IM3, 1, 0, 1, 66.63\\
P-0000625-T01-IM3, 1, 0, 0, 69.77\\
P-0000674-T01-IM3, 1, 0, 0, 65.33\\
\dots

\end{tcolorbox}

\begin{tcolorbox}[colback=lightboxgray, colframe=boxgray, boxrule=0.6pt, arc=3pt,
  left=8pt, right=8pt, top=4pt, bottom=4pt,
  title=\texttt{answer\_key.json} (hidden), fonttitle=\bfseries\small, coltitle=white, colbacktitle=boxgray]
\ttfamily\scriptsize
\{\\
\quad "test\_type": "Logistic regression",\\
\quad "treatment\_variable": "pten\_mutated",\\
\quad "treatment\_coefficient": -0.5059,\\
\quad "odds\_ratio": 0.603,\\
\quad "treatment\_p\_value": 0.0146,\\
\quad "significant\_at\_005": true,\\
\quad "conclusion": "Reject H0",\\
\quad "n\_observations": 2396\\
\}
\end{tcolorbox}

\subsection{Example from medicine}

\textit{Source:Knaus et al., ``The SUPPORT prognostic model. Objective estimates of survival for seriously ill hospitalized adults.'' Annals of Internal Medicine, 1995, 122:191--203 \citep{knaus1995support}} Prospective cohort of seriously ill hospitalized adults. 

\begin{tcolorbox}[colback=lightboxgray, colframe=boxgray, boxrule=0.6pt, arc=3pt,
  left=8pt, right=8pt, top=4pt, bottom=4pt,
  title=\texttt{analysis\_request.txt}, fonttitle=\bfseries\small, coltitle=white, colbacktitle=boxgray]
\small
\textbf{Research question:} Among seriously ill hospitalized adults in the SUPPORT study, is in-hospital mortality associated with patient disease class, particularly the Coma category?\\[3pt]
\textbf{Hypothesis:}\\
$H_0$: Patients in the Coma disease class do not differ in in-hospital mortality from the reference disease class.\\
$H_1$: Patients in the Coma disease class differ in in-hospital mortality from the reference disease class.\\[3pt]
\textbf{Study design:} Prospective cohort; 9{,}105 observations; reference analysis uses $N=5{,}732$ after task-specific filtering and missing-data handling.\\[3pt]
\textbf{Variable glossary:} \texttt{hospdead} (binary) [OUTCOME]; \texttt{dzclass} (categorical: ARF/MOSF w/Malignancy, ARF/MOSF w/Sepsis, COPD/CHF/Cirrhosis, Coma, Cancer) [TARGET]; \texttt{age}, \texttt{num.co}, \texttt{meanbp}, \texttt{hrt}, \texttt{resp}, \texttt{crea}, \texttt{alb} (covariates).
\end{tcolorbox}

\begin{tcolorbox}[
  colback=lightboxgray,
  colframe=boxgray,
  boxrule=0.6pt,
  arc=3pt,
  left=8pt,
  right=8pt,
  top=4pt,
  bottom=4pt,
  title={\ttfamily dataset.csv\ (first 3 rows, selected columns)},
  fonttitle=\bfseries\small,
  coltitle=white,
  colbacktitle=boxgray
]

\ttfamily\scriptsize

age, sex, hospdead, dzclass, num.co\\
62.85, male, 0, Cancer, 0\\
60.34, female, 1, COPD/CHF/Cirrhosis, 2\\
52.75, female, 0, COPD/CHF/Cirrhosis, 2\\
\dots

\end{tcolorbox}

\begin{tcolorbox}[colback=lightboxgray, colframe=boxgray, boxrule=0.6pt, arc=3pt,
  left=8pt, right=8pt, top=4pt, bottom=4pt,
  title=\texttt{answer\_key.json} (hidden), fonttitle=\bfseries\small, coltitle=white, colbacktitle=boxgray]
\ttfamily\scriptsize
\{\\
\quad "test\_type": "Logistic regression",\\
\quad "treatment\_variable": "dzclass (Coma)",\\
\quad "treatment\_coefficient": 0.9477,\\
\quad "treatment\_p\_value": 0.0001,\\
\quad "significant\_at\_005": true,\\
\quad "conclusion": "Reject H0",\\
\quad "n\_observations": 5732\\
\}
\end{tcolorbox}

\section{Fisher-R1 Training Details}

\subsection{Why R rather than Python}
\label{app:r-vs-python}

We use R rather than Python because the canonical implementations of the statistical methods P-Bench covers --- Cox proportional-hazards regression, mixed-effects models, instrumental-variable estimators, robust and rank-based tests --- are most mature and standardized in R packages such as \texttt{survival}, \texttt{lme4}, \texttt{rms}, and \texttt{AER}, whose outputs are widely treated as canonical in biomedicine, biostatistics, and economics. Python's statistical stack (\texttt{statsmodels}, \texttt{scipy.stats}, \texttt{lifelines}) is less complete on these methods and frequently differs in default settings (e.g., degrees-of-freedom corrections, link parameterizations, robust-variance estimators) that are standardized across R packages. Statisticians typically use R for these analyses, making it easier for domain experts to verify the agent's executions against the reference analysis in a single software ecosystem.

\subsection{Synthetic Data Generations}
\label{app:synthetic-data}
This appendix gives the full implementation of the synthetic-task pipeline summarised in \S\ref{sec:train-rl}.

\subsubsection{End-to-end pipeline}

For each cell of the Cartesian grid $M \times S_m \times N \times E \times P \times K$ in the task taxonomy, we construct one task in five stages.

\paragraph{Stage 1: simulation template.}
For each statistical method $M$, we use an LLM to draft a parameterised simulation script. The script implements a canonical data-generating process for that method (e.g., a two-stage least-squares DGP for IV with an instrument, an endogenous treatment, an outcome, and covariates) and exposes the relevant parameters as variables (sample size, effect size, error variance, clustering, fixed effects, etc.). All templates were drafted by Claude Sonnet 4.6 and checked by the authors to fix syntax errors and standardise the parameter-slot naming. Templates are written once per method and reused across all tasks for that method.

\paragraph{Stage 2: parameter instantiation.}
Each cell of the grid specifies a sample size, an effect size, and a seed. We plug these into the simulation template to fix all parameter values. 

\paragraph{Stage 3: execution and answer key.}
We execute the instantiated script. It writes a CSV dataset and runs the canonical statistical method on that dataset. The reported $p$-value, the reject/fail-to-reject decision at $\alpha=0.05$, and the canonical method label are recorded in \texttt{answer\_key.json}. Crucially, the answer key is the \emph{output of the canonical method on the simulated data}, not the true generative parameter. This aligns the supervision target with what a correct analysis would actually produce.

\paragraph{Stage 4: question rendering.}
Each scenario provides two prompt templates: \texttt{analysis\_nohint} states only the research question, while \texttt{analysis\_hint} adds a one-sentence study-design summary (see Table~\ref{tab:hint-examples}). The chosen prompt style $P$ selects which template is rendered, and the result is written to \texttt{analysis\_request.txt}.

\paragraph{Stage 5: task package.}
Each task consists of four files: \texttt{study\_description.txt}, \texttt{analysis\_request.txt}, \texttt{dataset.csv}, and \texttt{answer\_key.json}. The agent sees the first three. The answer key is hidden and used only for scoring.

\subsection{Per-axis level lists}

\begin{table}[h]
\centering
\small
\renewcommand{\arraystretch}{1.15}
\caption{Levels for each axis of the synthetic-task Cartesian grid.}
\label{tab:synthetic-axes}
\begin{tabular}{l p{0.8\linewidth}}
\toprule
\textbf{Axis} & \textbf{Levels} \\
\midrule
$M$ method & 27 statistical methods spanning parametric, rank-based, regression, causal, hierarchical, and survival families \\
$S_m$ scenario & 5--7 domain scenarios per method. Each scenario specifies a column-renaming map, a study description, and \texttt{hint}/\texttt{non-hint} prompt templates. \\
$N$ sample size & small, medium, large, xlarge. Exact values are method-specific; across methods, $N$ ranges from $\sim$30 for paired and small two-sample tests, $\sim$200--500 for typical regressions, up to $\sim$4{,}800 for hierarchical and panel models. \\
$E$ effect size & null, borderline, medium. \\
$P$ prompt style & \texttt{hint}, \texttt{non-hint}. See \ref{tab:hint-examples} for examples. \\
$K$ seed & $s_0$, $s_1$, $s_2$. Three deterministic seeds per cell. \\
\bottomrule
\end{tabular}
\end{table}

\begin{table}[h]
\centering
\small
\renewcommand{\arraystretch}{1.2}
\caption{Examples of \texttt{non-hint} and \texttt{hint} prompts across method types. Both variants ask the same research question; the \texttt{hint} variant adds a one-sentence study-design summary that flags a methodologically relevant feature.}
\label{tab:hint-examples}
\begin{tabular}{p{0.18\linewidth} p{0.38\linewidth} p{0.38\linewidth}}
\toprule
\textbf{Method} & \textbf{\texttt{non-hint}} & \textbf{\texttt{hint}} \\
\midrule
Mann--Whitney &
``Does API latency differ between server A and server B?'' &
``\ldots{} The latency distribution is heavily right-skewed.'' \\
Mixed-effects &
``Does the new ad creative improve user click-through rate?'' &
``\ldots{} Each user is shown multiple ads over time, with users nested within demographic segments.'' \\
Cox PH &
``Does the new battery formulation extend laptop battery life?'' &
``\ldots{} Devices still in use at study end have right-censored lifetimes.'' \\
\bottomrule
\end{tabular}
\end{table}

\subsubsection{Example}

We give a single fully worked synthetic task to illustrate the pipeline end-to-end. The task uses the Mann--Whitney scenario \texttt{app\_latency\_two\_servers}, with cell values $N=$ medium ($n=200$), $E=$ medium, $K=s_0$, and prompt style $P=$ \texttt{hint}. The four files generated by the pipeline are shown below.

\begin{tcolorbox}[colback=lightboxgray, colframe=boxgray, boxrule=0.6pt,
  arc=3pt, left=8pt, right=8pt, top=4pt, bottom=4pt,
  title=\texttt{study\_description.txt}, fonttitle=\bfseries\small, coltitle=white, colbacktitle=boxgray]
\small
Paper: ``P50 vs P95 App Request Latency: Two-Pool A/B Comparison''\\[2pt]
Study: Independent web requests were routed to one of two server pools. \\[4pt]
Variables: \texttt{request\_id}, \texttt{server\_pool} (0 = pool A, 1 = pool B) [GROUP],
\texttt{request\_latency\_ms} (continuous, heavy right-tailed) [OUTCOME].\\[2pt]
Sample: 200 independent requests.
\end{tcolorbox}

\begin{tcolorbox}[colback=lightboxgray, colframe=boxgray, boxrule=0.6pt,
  arc=3pt, left=8pt, right=8pt, top=4pt, bottom=4pt,
  title=\texttt{analysis\_request.txt}, fonttitle=\bfseries\small, coltitle=white, colbacktitle=boxgray]
\small
Research question: Do the two server pools have different latency distributions?\\[4pt]
\\[2pt]
Report test statistic, $p$-value, location shift with 95\% CI,
conclusion at $\alpha = 0.05$.
\end{tcolorbox}

\begin{tcolorbox}[colback=lightboxgray, colframe=boxgray, boxrule=0.6pt,
  arc=3pt, left=8pt, right=8pt, top=4pt, bottom=4pt,
  title=\texttt{dataset.csv} (first 6 of 200 rows),
  fonttitle=\bfseries\small, coltitle=white, colbacktitle=boxgray]
\ttfamily\scriptsize
request\_id,server\_pool,request\_latency\_ms\\
1,0,87.32\\
2,1,142.05\\
3,0,71.48\\
4,1,239.71\\
5,0,103.86\\
6,1,118.40\\
\dots
\end{tcolorbox}

\begin{tcolorbox}[colback=lightboxgray, colframe=boxgray, boxrule=0.6pt,
  arc=3pt, left=8pt, right=8pt, top=4pt, bottom=4pt,
  title=\texttt{answer\_key.json} (hidden from agent),
  fonttitle=\bfseries\small, coltitle=white, colbacktitle=boxgray]
\ttfamily\scriptsize
\{\\
\quad "task\_id":           "task\_app\_latency\_two\_servers\_medium\_medium\_hint\_s0",\\
\quad "method":            "mann\_whitney\_test",\\
\quad "test\_statistic":   4521.0,\\
\quad "p\_value\_correct": 0.0184,\\
\quad "location\_shift":   12.4,\\
\quad "ci\_95":             [3.1, 24.7],\\
\quad "conclusion":        "Reject H0",\\
\quad "n\_observations":  200,\\
\quad "naive\_method":    "Welch's t-test",\\
\quad "p\_value\_naive":  0.0712\\
\}
\end{tcolorbox}

The agent sees only the first three files. The correct analysis is a Mann–Whitney test, which yields $p \approx 0.018$ and rejects $H_0$. A naive Welch's $t$-test would yield $p \approx 0.07$ and reach the wrong conclusion.

\begin{figure}[h]
\centering
\includegraphics[width=0.5\textwidth]{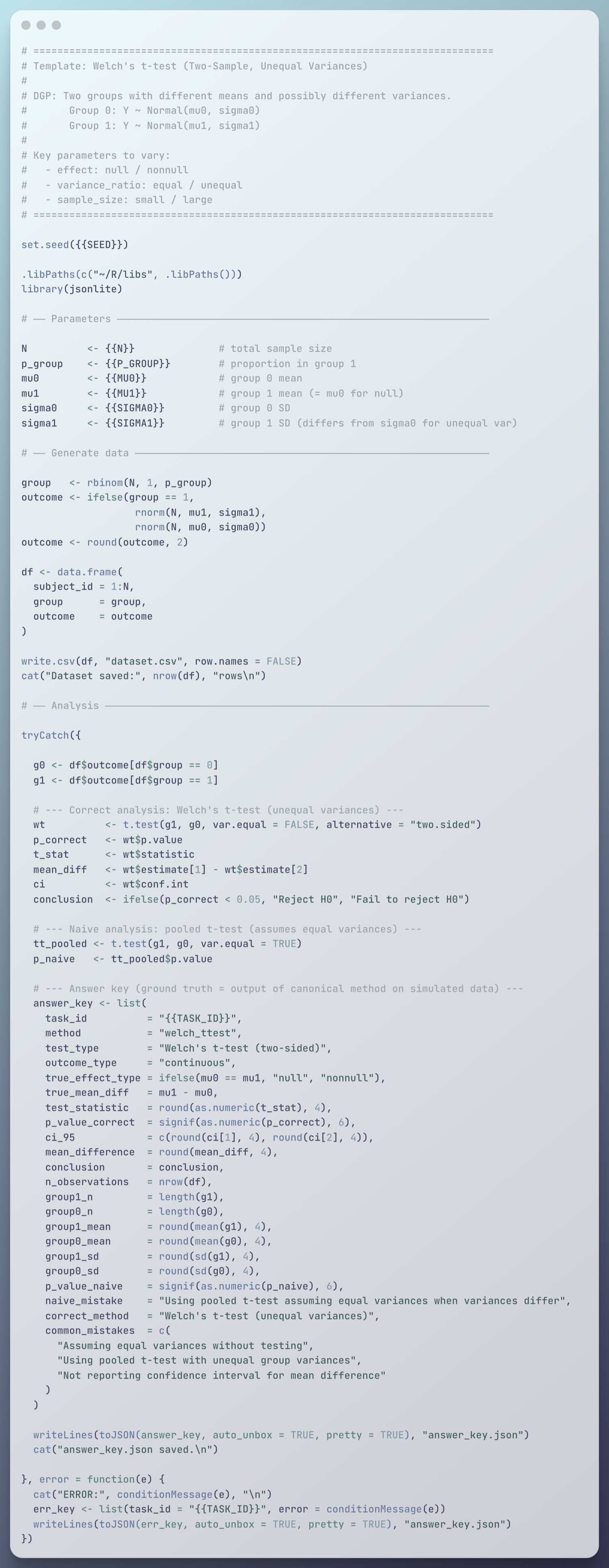}
\caption{Full Welch's t-test simulation template.}
\label{lst:welch-template}
\end{figure}

\subsubsection{Data-quality perturbations}

On top of the clean Cartesian grid, we add 60 perturbed task variants drawn from regression-method cells. The two perturbation types are: (i) \texttt{missing} --- a fraction of values in the outcome or a covariate are set to \texttt{NA}, requiring the agent to choose how to handle missingness; (ii) \texttt{outlier} --- a small number of rows are injected with extreme or invalid values (e.g., physiologically implausible measurements), requiring the agent to detect and address them before fitting. The answer key for each perturbed variant is computed by running the canonical method on the perturbed data with the standard handling protocol for the issue (e.g., listwise deletion for \texttt{missing}, robust standard errors or trimming for \texttt{outlier}).

\subsection{Training setup and detailed hyperparameters}
\label{app:hyperparameters}
We conduct SFT using LlamaFactory~\citep{zheng2024llamafactoryunifiedefficientfinetuning} and RL training using \texttt{verl}~\citep{sheng2025hybridflow}. For SFT, we train for 3 epochs with a learning rate of $1\mathrm{e}{-5}$, a cosine learning-rate scheduler, a warmup ratio of $0.1$, a batch size of $16$, and a cutoff length of $8192$ tokens. For RL, we train for 1 epoch with a learning rate of $1\mathrm{e}{-6}$, constant learning-rate warmup, and $20$ warmup steps. The batch size is $16$, the maximum prompt length is $2048$ tokens, and the maximum response length is $4096$ tokens. We use asymmetric clipping with $\epsilon_{\mathrm{low}}=0.2$ and $\epsilon_{\mathrm{high}}=0.28$. During rollout generation, we use temperature $0.7$, top-$p=1.0$, and a rollout group size of $G=8$. At inference time, we use temperature $0.3$, top-$p=0.9$, and a batch size of $10$.

\subsection{Computing Resources}
All training and evaluation runs are conducted on a single node with NVIDIA H200 GPUs
(141 GB HBM each).  For the 14B model, the full RL training run completes in about 36 hours.
\subsection{More Experiment Results}
\begin{table}[h]
\centering
\caption{Results on P-Bench under the relaxed strict threshold. Raw denotes correct conclusion direction, while Strict additionally requires $|\Delta z| < 1.0$, where $z$ is the two-sided $z$-score derived from the reported $p$-value. pass@1 is reported as mean$_{\pm \mathrm{stdev}}$ over three independent runs.}
\label{tab:main_results_z1}
\small
\setlength{\tabcolsep}{5pt}
\renewcommand{\arraystretch}{1.15}

\begin{adjustbox}{max width=\linewidth}
\begin{tabular}{l*{8}{c}}
\toprule
\textbf{Model}
& \multicolumn{4}{c}{\textbf{P-Easy}}
& \multicolumn{4}{c}{\textbf{P-Hard}} \\
\cmidrule(lr){2-5} \cmidrule(lr){6-9}
& \multicolumn{2}{c}{Raw}
& \multicolumn{2}{c}{Strict}
& \multicolumn{2}{c}{Raw}
& \multicolumn{2}{c}{Strict} \\
\cmidrule(lr){2-3} \cmidrule(lr){4-5}
\cmidrule(lr){6-7} \cmidrule(lr){8-9}
& pass@1 & pass@3
& pass@1 & pass@3
& pass@1 & pass@3
& pass@1 & pass@3 \\
\midrule

\multicolumn{9}{c}{\textit{Proprietary Models}} \\
\midrule
GPT-5.4
& $\mathbf{92.9}_{\pm 1.0}$ & 95.6
& $\mathbf{72.6}_{\pm 2.5}$ & 79.3
& $58.3_{\pm 0.9}$ & 68.0
& $38.4_{\pm 0.7}$ & 46.0 \\

\midrule
\multicolumn{9}{c}{\textit{Open-source Models}} \\
\midrule
DeepSeek V4 Pro
& $70.9_{\pm 0.5}$ & 93.1
& $59.1_{\pm 0.9}$ & 78.8
& $46.8_{\pm 1.2}$ & 69.7
& $32.2_{\pm 1.1}$ & 49.3 \\

GPT-OSS-120B
& $48.1_{\pm 4.3}$ & 86.2
& $31.2_{\pm 4.4}$ & 64.5
& $37.4_{\pm 4.3}$ & 71.2
& $23.6_{\pm 2.3}$ & 49.1 \\

Qwen-3-Coder-30B
& $76.9_{\pm 2.2}$ & 87.7
& $53.2_{\pm 1.7}$ & 65.0
& $59.8_{\pm 0.9}$ & 68.0
& $35.4_{\pm 1.0}$ & 42.3 \\

Qwen-3-32B
& $82.8_{\pm 0.0}$ & 90.6
& $55.2_{\pm 0.5}$ & 66.0
& $58.4_{\pm 0.7}$ & 71.6
& $34.2_{\pm 0.5}$ & 46.4 \\

Qwen-2.5-Coder-7B
& $61.4_{\pm 8.5}$ & 83.7
& $41.4_{\pm 6.5}$ & 60.1
& $37.5_{\pm 2.8}$ & 63.1
& $18.6_{\pm 0.7}$ & 32.0 \\

DataMind-7B
& $60.3_{\pm 1.0}$ & 84.2
& $38.6_{\pm 1.4}$ & 57.1
& $44.4_{\pm 4.2}$ & 66.7
& $25.2_{\pm 2.1}$ & 40.5 \\

\rowcolor{orange!14}
\textbf{Fisher-R1-7B}
& $87.0_{\pm 1.6}$ & $\mathbf{96.6}$
& $71.8_{\pm 0.8}$ & $\mathbf{80.8}$
& $63.4_{\pm 2.5}$ & $\mathbf{82.4}$
& $40.1_{\pm 2.5}$ & $\mathbf{56.8}$ \\

Qwen-2.5-Coder-14B
& $78.5_{\pm 1.2}$ & 88.2
& $50.3_{\pm 1.3}$ & 58.1
& $51.5_{\pm 4.3}$ & 66.7
& $30.8_{\pm 2.1}$ & 41.0 \\

DataMind-14B
& $69.0_{\pm 5.2}$ & 88.2
& $47.8_{\pm 3.7}$ & 64.0
& $48.2_{\pm 2.1}$ & 67.1
& $29.0_{\pm 0.9}$ & 42.3 \\

\rowcolor{orange!14}
\textbf{Fisher-R1-14B}
& $87.4_{\pm 1.9}$ & 94.1
& $70.6_{\pm 1.4}$ & 79.8
& $\mathbf{65.8}_{\pm 0.8}$ & 81.1
& $\mathbf{42.0}_{\pm 2.2}$ & $\mathbf{56.8}$ \\

\bottomrule
\end{tabular}
\end{adjustbox}
\end{table}

\section{Prompt}
\begin{promptbox}{Training Prompt}
\textbf{\# System Prompt}

You are a statistician working through a hypothesis-testing task in a multi-turn loop. The dataset is pre-loaded as `data` (a data.frame). Do not load it yourself.
On every turn, write exactly one of the following structures and then STOP:
\medskip

1. An intermediate analysis turn:

<think>

your reasoning about the data, assumptions, and the next analysis step

</think>
\par\medskip\noindent
<code>

one block of R code (use cat() or print() to display results)

</code>

\medskip

2. A final answer turn (no code):

<think>

brief summary of what you learned

</think>
\par\medskip\noindent
<answer>

1. Analysis summary: key EDA findings, assumption check results, method choice and rationale.
\medskip

2. Results interpretation: effect estimate, direction, magnitude, p-value, confidence interval.

\medskip
FINAL ANSWER:

- Treatment effect: [number]

- P-value: [number]

- 95\% CI: [lower, upper]

- Conclusion: [Reject H0 / Fail to reject H0] at alpha = 0.05

</answer>

\medskip

Rules:

- After each <code> block, the runtime will reply with <observation>...</observation> containing the R stdout. Read it before writing the next turn.

- Do NOT write more than one <code> block per turn. Do NOT write <observation> yourself.

- Do NOT print raw data values, unique values, or large vectors. Use dim(), colnames(), and class() to inspect data.

- If your code produces an error, read the message carefully and fix the root cause before retrying.

- Only emit <answer> on the very last turn, and only after you have verified the numerical results from <observation>.

\bigskip
\textbf{\# User Prompt}

\par\vspace{0.6em}\noindent
\#\# Research Question

\noindent\texttt{\{analysis\_request\}}

\par\vspace{0.6em}\noindent
\#\# Dataset

\noindent The dataset has been pre-loaded as \texttt{data}.

\end{promptbox}
\newpage

\end{document}